%% file: acl2019.tex
\documentclass[11pt,a4paper]{article}
\usepackage[hyperref]{acl2019}
\usepackage{times}
\usepackage{latexsym}
\usepackage{url}
\usepackage{xcolor}
\newcommand{\ignore}[1]{}

\newcommand{\rn}{Distribution over Quantities} 
\newcommand{\rnacronym}{\textsc{DoQ}}

\newcommand{\fc}{\textsc{Orig F\&C}}
\newcommand{\fcc}{\textsc{Clean F\&C}}
\newcommand{\fcl}{\textsc{No-Leak F\&C}}

\usepackage{multirow} 
\usepackage{graphicx} 
\usepackage{lingmacros}
\usepackage{algorithm} 
\usepackage{algpseudocode}
\usepackage{adjustbox} 
\usepackage{subfig}
\usepackage{multicol}
\usepackage{textcomp} 

\aclfinalcopy 

\title{How Large Are Lions? \\ 
Inducing Distributions over Quantitative Attributes}

\author{
  Yanai Elazar\thanks{~~Work carried out during an internship at Google.} \\
  Bar Ilan Univesity \\
    \texttt{\normalsize yanaiela@gmail.com} \\
\And
  Abhijit Mahabal\thanks{~~Work carried out during employment at Google.} \\
  Pinterest \\
  \texttt{\normalsize amahabal@gmail.com}
\AND Deepak Ramachandran \\
  Google Research\\
   \texttt{\normalsize ramachandrand@google.com} \\
\And
  Tania Bedrax-Weiss \\
  Google Research\\
  \texttt{\normalsize tbedrax@google.com} \\
\And
  Dan Roth  \\
  University of Pennsylvania \\
   \texttt{\normalsize danroth@seas.upenn.edu} \\
}
\date{}

\begin{document}
\maketitle

\begin{abstract}
Most current NLP systems have little knowledge about  quantitative attributes of objects and events. We propose an unsupervised method for collecting quantitative information from large amounts of web data, and use it to create a new, very large resource consisting of distributions over physical quantities associated with objects, adjectives, and verbs which we call \rn\ (\rnacronym)\footnote{The resource is available at \url{https://github.com/google-research-datasets/distribution-over-quantities}}.
This contrasts with recent work in this area which has focused on making only relative comparisons such as ``Is a lion bigger than a wolf?".
Our evaluation shows that \rnacronym\ compares favorably with state of the art results on existing datasets for relative comparisons of nouns and adjectives, and on a new dataset we introduce.
\end{abstract}

\input{intro.tex}

\input{related.tex}

\input{resource.tex}

\input{eval.tex}

\input{results.tex}

\input{discussion.tex}

\section*{Acknowledgments}
We would like to thank Ellie Pavlick, Jason Baldridge, Anne Cocos, Vered Shwartz, Hila Gonen and the 3 anonymous reviewers for helpful comments. Furthermore, we thank Maxwell Forbes, Yiben Yang and Niket Tandon for their helpful clarifications regarding their methods and code.
The research of Dan Roth is partly supported by a Google gift and by DARPA, under agreement number HR0011-18-2-0052.
\bibliography{acl2019}
\bibliographystyle{acl_natbib}

\newpage
\clearpage
\appendix

\input{supplemental.tex}

\end{document}

%% file: intro.tex
\section{Introduction}

How much does a lion weigh? How tall can they be? When do people typically eat breakfast? And, how long are concerts?
Most people would know at least an approximate answer to these questions, many of which fall under the (somewhat ill-defined) notion of commonsense knowledge, and some (but certainly not all) of which exist in resources such as Wikipedia, or in knowledge graphs like Freebase \cite{bollacker2008freebase}.
Natural Language Understanding systems should also know (at least approximately) the answers to these questions, to better support Question Answering and Textual Entailment~\cite{DRSZ13} and, more generally, in order to support reasoning about events described in natural language  and converse with people naturally.

\begin{figure}[t!]
\centering
\subfloat[Mass distributions for multiple animals.]{
\includegraphics[width=1.0\columnwidth]{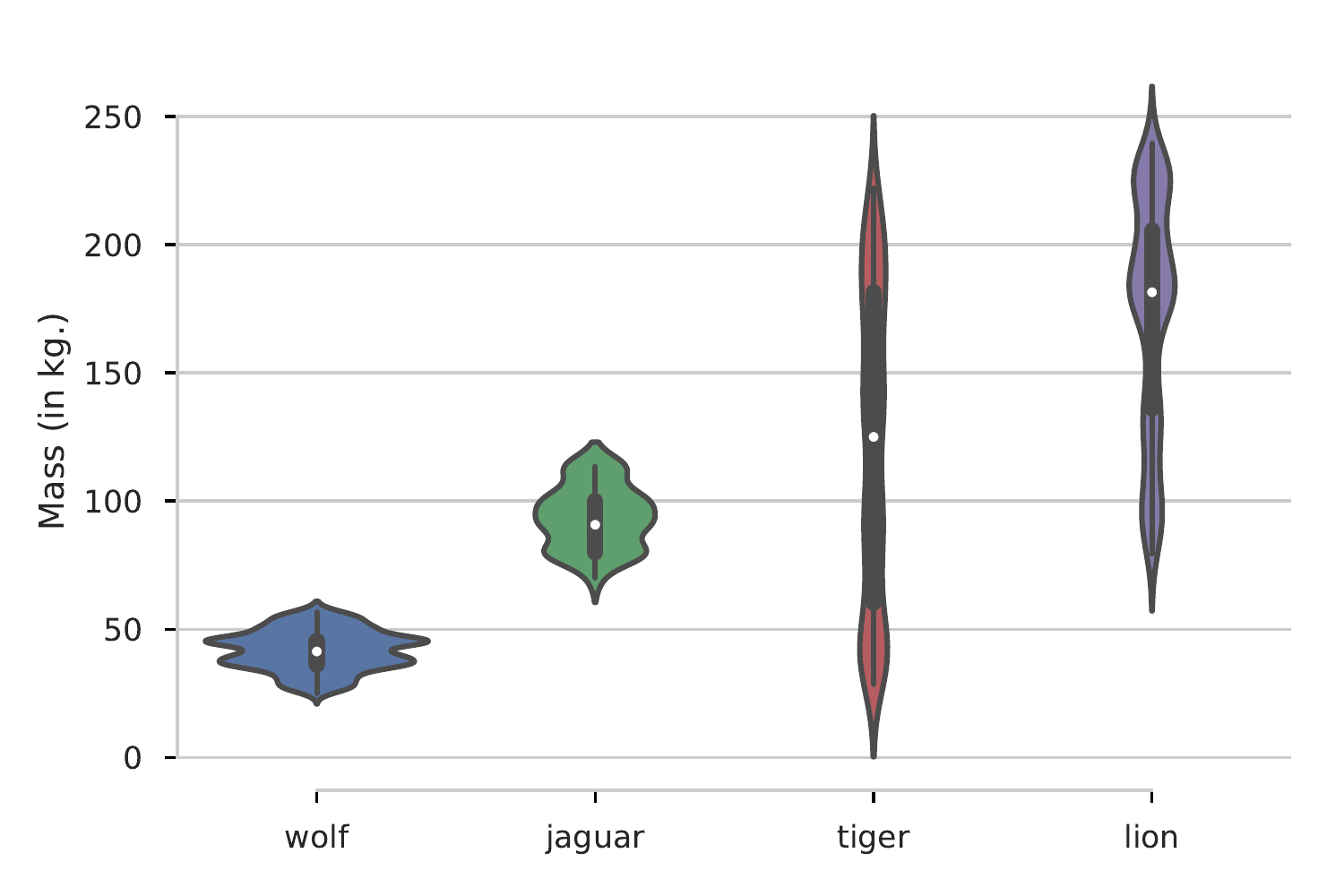}
\label{fig:animal-mass}
}\\

\centering
\subfloat[Typical hours for different meals of the day.]{
\includegraphics[width=1.0\columnwidth]{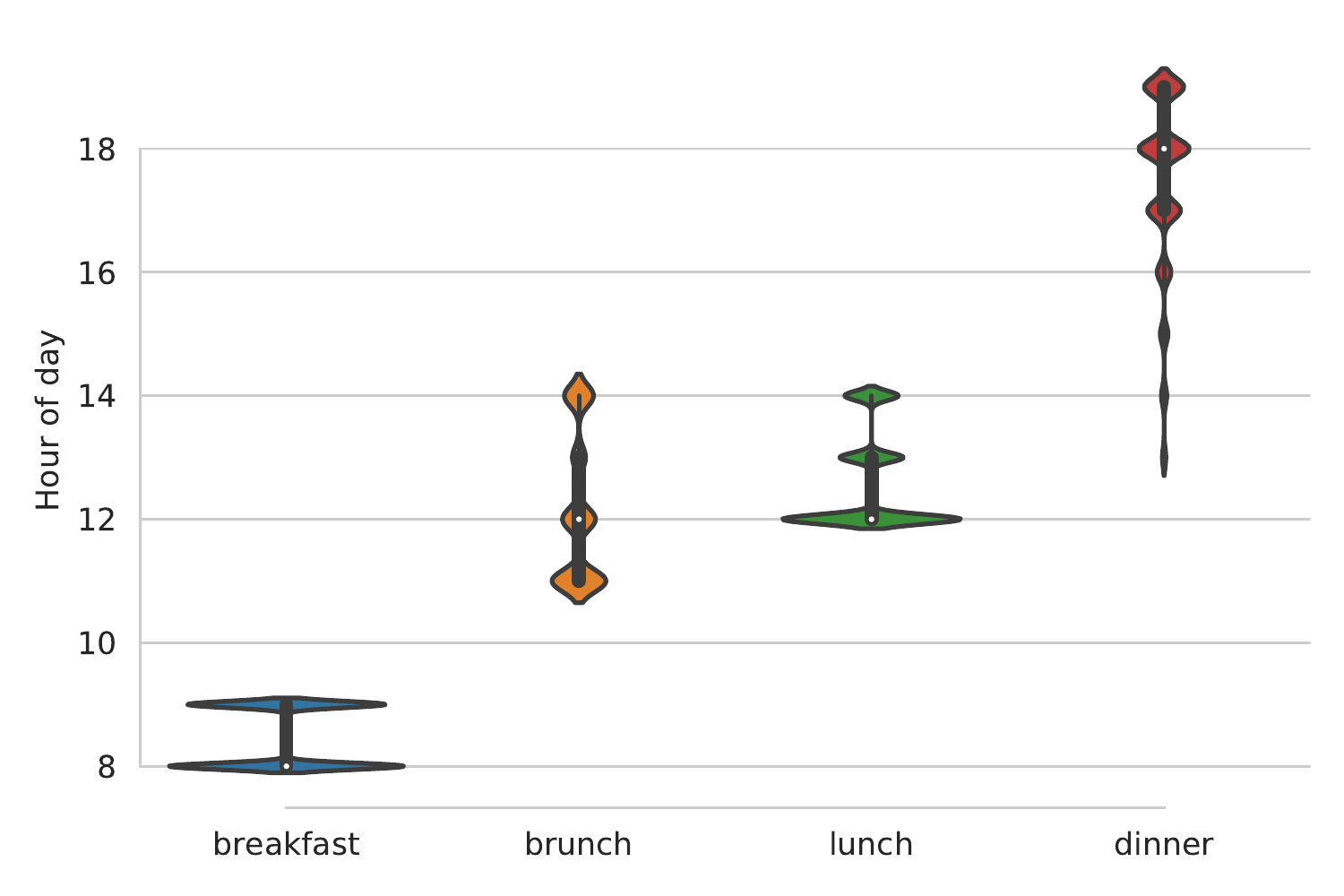}
\label{fig:eating-hours}
}

\caption{Examples of different objects from \rnacronym\ in the \textsc{mass} and \textsc{time} dimensions.}
\label{fig:DoQ_examples}
\end{figure}

Acquiring commonsense knowledge from natural language text has been the subject of a lot of recent work.
These approaches focus on facilitating comparisons between quantitative attributes of nouns~\cite{bagherinezhad2016elephants,forbes2017verb,yang2018extracting}, intensity of adjectives \cite{de2013good,D18-1202} and  coarse classification of events duration and relative order \cite{gusev2011using,NWPR18}. However, they do not have complete coverage even for comparable objects, as a result of how they are acquired, and lack the ability to assign a numerical value to objects and events (``How hot is it in New York?"), which is often useful for reasoning, text generation, and other tasks.

In this work, we propose a method for acquiring distributions over ten dimensions: \textsc{time}, \textsc{currency}, \textsc{length}, \textsc{area}, \textsc{volume}, \textsc{mass}, \textsc{temperature}, \textsc{duration}, \textsc{speed}, and \textsc{voltage}. We do this for nouns (e.g. elephant, airplane, NBA game), adjectives (e.g. cold, hot, lukewarm) and verbs (e.g. eating, walking, running). This results in a large resource we call \rn\ (\rnacronym)\ -- over $350K$ triples each observed over $1000$ times.
Examples of entries in \rnacronym\ depicting \textsc{mass} and \textsc{time} distributions are shown in Figures \ref{fig:animal-mass} and \ref{fig:eating-hours}.\footnote{The violin plots along the paper describe the probability density of the collected distribution at different values.}

We develop \rnacronym\ by extracting and aggregating quantitative information from the web, in English, and collecting co-occurring objects from their surroundings. The quantitative information is normalized and associated with units to determine the relevant dimensions such as \textsc{temperature} or \textsc{mass}. 
As we show, despite the inherent noise in such an acquisition process due to extraction errors and reporting bias \cite{gordon2013reporting}, rather simple denoising methods result in a relatively clean resource, with very high coverage and good accuracy. 

\rnacronym\ is significantly more comprehensive and accurate than any other related resource we know of. For each term, and each of its relevant dimensions we collected the actual numerical values associated with this pair. This gives us expressive distributional information about range, mean, median and other statistics. Moreover, since our resource is collected using only a few rules for detecting quantities and converting units, it can be extended to other languages easily.

We evaluate \rnacronym\ on several existing datasets and show that it compares favorably with existing methods that require more resources and have less coverage. In particular, we identify and correct problems with some of the existing datasets resulting in new, cleaner, evaluation datasets.  

Overall, we make the following contributions: 

\begin{enumerate}
    \item A new method for collecting expressive quantitative information about objects.
    \item A large resource of distributions over quantitative attributes of nouns, adjectives, and verbs.
    \item Strong results on existing datasets for noun and adjective comparison, refining and improving an existing dataset, and creating a new dataset for evaluating noun comparisons.
\end{enumerate}

%% file: related.tex
\section{Related Work}

There has been a lot of work trying to use Hearst-style patterns \cite{hearst1992automatic} to extract relations between objects in large corpora \cite{tandon2014acquiring,shivade2016identification}. For example, from the sentence: \textit{``Melons are bigger than apples''} they extract the  relation: `\textit{Melons}' $>$ `\textit{apples}'.
These methods suffer from reporting bias and low coverage, since the precise patterns need to be found to make these inferences. Our method, which relies on co-occurring objects, is robust to this issue. 
Pattern-based methods were also used in the context of OpenIE, e.g., to extract event duration information \cite{gusev2011using,kozareva2011learning}, but were found to be highly brittle due to the dependence on finding specific pre-defined patterns.

There is a line of work \cite{forbes2017verb,yang2018extracting} to  determine the quantitative relation between two nouns on a specific scale.
For adjectives~\cite{de2013good,kim2013deriving,shivade2015corpus,D18-1202}, comparisons were made only for relative intensities, i.e. `\textit{freezing}' $<$ `\textit{cold}'. 
In contrast, we infer magnitudes as well, which make us robust to comparisons between different polarities of the same cluster (e.g. `\textit{hot}' vs. `\textit{cold}').


 \citet{spithourakis2018numeracy} propose several methods to represent numbers in language models (LMs) instead of using an out-of-vocabulary token, giving the LM more expressive ability to produce numbers. \citet{spithourakis2016numerically} showed that conditioning on numerical values in the LM can improve the consistency of the modeling for clinical reports. When using it along with a scorer for Semantic Error Correction \cite{dahlmeier2011correcting}, it makes more grounded suggestions, with realistic estimates of different measurements.

Our work overlaps with a number of approaches to ground textual objects by: achieving a commonsense understanding of numeric expressions \cite{chaganty2016much}, grounding adjectives into RGB colors \cite{winn2018lighter}, grounding events duration \cite{pan2006annotated,gusev2011using} and measurements' intensity within a given context \cite{narisawa2013204}.

Finally, our resource collection is in the line of work that uses counting across very large amounts of data (such as n-grams from books) to produce big resources \cite{lin2012syntactic,goldberg2013dataset}, which have had a significant impact on NLP Research.

%% file: resource.tex
\section{\rn: Method}

\begin{figure}[t]
\centering
\includegraphics[width=1.0\columnwidth]{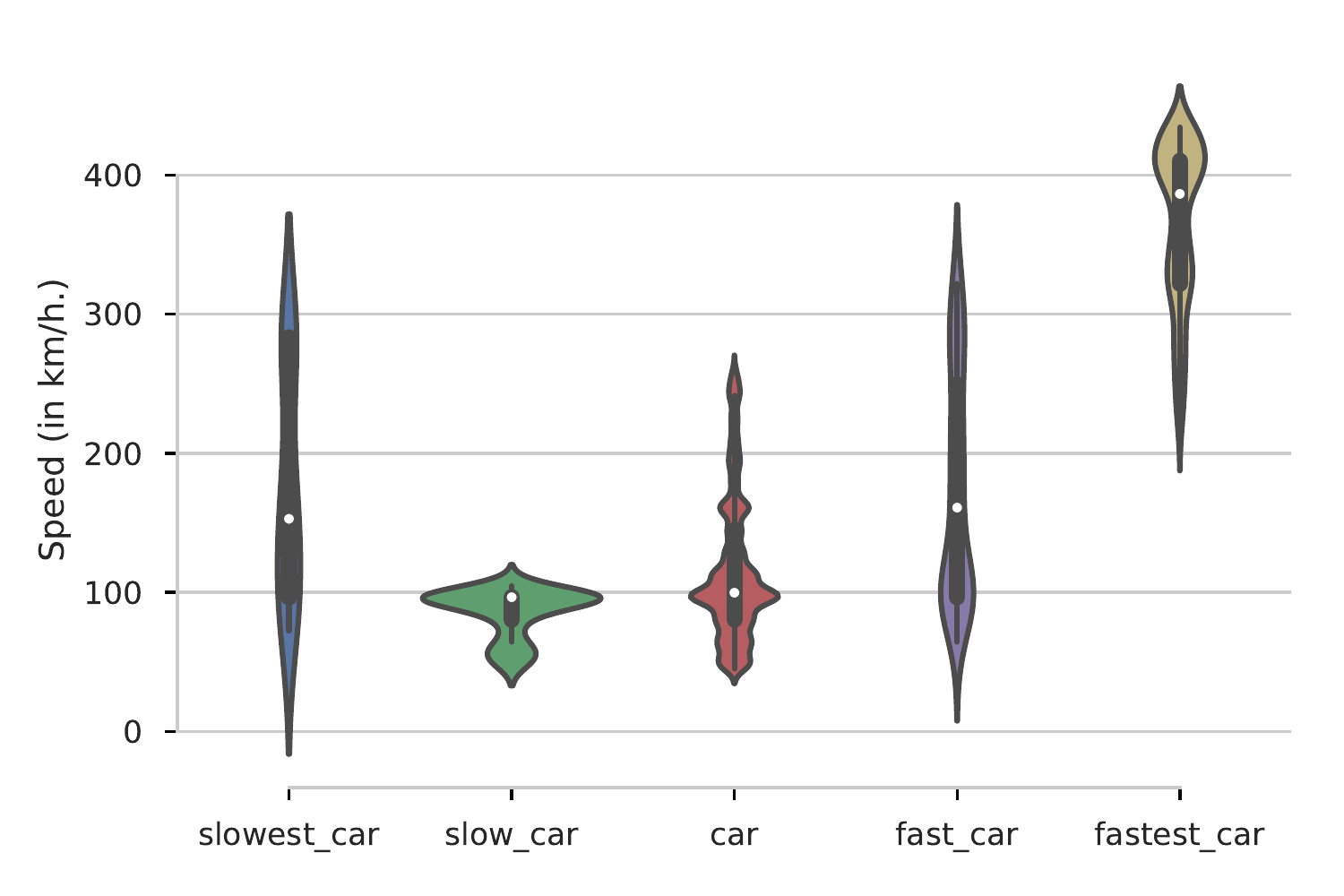}
\caption{Car modifiers: Speed of cars, sidelong by different modifiers, which shift the cars speed distribution. Interesting to point out the high distribution of ``slowest car'' phrase (See the bias discussion in Sec. \ref{sec:discussion}).}
\label{fig:adj-comparison}
\end{figure}

We propose a process for automatically extracting co-occurrences of objects and measurements from a large text corpus. Examples of the resulting output are the mass distributions of animals in Figure \ref{fig:animal-mass}, typical meal hours in Figure \ref{fig:eating-hours} and the car modifiers in Figure \ref{fig:adj-comparison}.

We first use a rule-based method for detecting and normalizing measurement mentions (Sec. \ref{sec:measure-identify}).
We then aggregate the detected measurements and objects that occurred in the nearby context (Sec. \ref{sec:obj-collection}) and describe some simple heuristics for improving the resource accuracy (Sec. \ref{sec:denoising}).
Finally, Sec. \ref{sec:resource} describes the resource produced in this process.

We note that the resource was built with the aim of keeping it as simple as possible, to test how accurate a simple approach can be. We believe it reflects the potential of transferring the process to other languages, where NLP resources are more sparse.

\subsection{Measurement Identification and Normalization}
\label{sec:measure-identify}
Measurement identification uses a simple context-free grammar along with a mapping from units to dimensions. Thus, we know that \textit{`inch'} is a unit in the \textsc{length} domain which is equal to 0.02524 meters, and that ``acre foot" is a unit of \textsc{volume} equal to 1233.48 standard units (here, cubic meters). Similar tables express \textsc{speed} in meters per second and \textsc{temperature} in degree Kelvin.

If the unit is not expressed explicitly or recognized by the parser (for example, in the sentence ``\textit{New York was a scorching 110}''), we do not extract anything. There are occasional mis-parses caused by typographic shortcuts, such as ``\textit{17 C}'' where Centigrade is meant, but is parsed as Coulombs. These show up as loss in coverage for us, since we deal with a limited set of dimensions in which charge is not included.


\subsection{Object Collection}
\label{sec:obj-collection}

\paragraph{Object Extraction}
The main objects used in this work are 1-token words that are either nouns, adjectives or verbs. We also consider more complex phrases of these types (e.g. noun phrases).
The complex phrases are collected enforcing minimum phrase spans. This way, for example, we collected the phrase ``\textit{race car}'' and are able to compare its distribution to that of ``\textit{electric car}''.

\paragraph{Object Head}
Along with each collected object, we also retrieve its syntactic head. For example, in the sentence: ``\textit{The fast car was driving 50 miles per hour}'', collecting the adjective `\textit{fast}' will also capture `\textit{car}' as its head. With this information we are able to compare a ``\textit{fast car}'' to a `\textit{car}'.
We note that this process is not possible for all languages and may result in less accurate extraction depending on the parser accuracy. Nonetheless, this phase is optional as it only adds the ability to compare more complex phrases and modifiers. A lot of information can still be collected without it.

\paragraph{Aggregation}
After identifying measurements in the sentence, we collect the objects that co-occur with these measurements within a certain context window. Using a bigger context size, we get broader coverage but also fewer occurrences. When reducing the context size, we get a sparser resource, but better attribution accuracy.
More sophisticated collection methods are possible (e.g. measuring parse-tree distances), but are left for future work.

\paragraph{Running the Entire Process}
We created the \rnacronym\ resource using the Flume framework \cite{Chambers:2010:FEE:1806596.1806638}, to quickly processes billions of English webpages in parallel. First, we identified and normalized measurements (Sec.~\ref{sec:measure-identify}). Then, these sentences were parsed for POS tags and dependency trees \cite{andor2016globally} and the relevant objects gathered by identifying co-occurences (within sentence or distance threshold). The following step aggregated all of the objects with the same object-head-measurement tuple, creating a distribution of numbers (Sec.  \ref{sec:obj-collection}).

\subsection{De-noising}
\label{sec:denoising}
The output of the described resource collection process is, as expected, quite noisy. It assumes a very simplified model of language, where co-occurring objects and numerical measurement are assumed equivalent to attribution, ignoring negations and reporting bias \cite{gordon2013reporting}. To address this, we employ de-noising filters focused on increasing precision.
We get a cleaner resource at the expense of coverage, which is still valuable due to the high volume of data used.

\paragraph{Distance Based Co-Occurrences}
When aggregating co-occurrences, we also record the token distance between the measurement and the object. This can be a good indication of the degree of relatedness of a word to its surroundings.
We used two context distances in our experiments: (1) co-occurrence within the same sentence, (2) co-occurrence within a token distance $k$.\footnote{In practice, we use $k=3,10$.}
In our experiments, we explore the effectiveness of the resource with different distance thresholds.

\paragraph{Negation}
Negations can affect the precision of the resource and contribute a lot to the distribution tails, as in: ``\textit{The dimension of the \textbf{car} is \textit{not} \underline{50cm}.}"  We decided to simply discard all  measurements that appear in the same sentence with a negation word.\footnote{Specifically, we used the following negation words: `not', `no', `without', `neither' and `nor'.}



\subsection{\rn\ Statistics}
\label{sec:resource}

The final resource contains 117,953,900 unique noun tuples, 2,513,033 unique adjective tuples and 2,121,448 unique verb tuples. The total number of triples in English are 122,588,381. Table \ref{tbl:resource-filter-pos-counts} provides some more statistics.

\input{tables/resource_stats.tex}

%% file: tables/resource_stats.tex
\begin{table}[t]
\resizebox{\columnwidth}{!}{%
\begin{tabular}{l|rrr}
Filter/Type & Nouns         & Adjectives & Verbs \\ \hline
none        & 117,953,900   & 2,513,033  & 2,121,448      \\
5           & 16,188,215    & 598,563    & 603,799      \\
100         & 1,497,753     & 130,534    & 160,060      \\
1000        & 266,655       & 40,518     & 51,625    
\end{tabular}
}
\caption{{\bf Size:} Number of tuples of Nouns, Adjectives, and Verbs, coupled with dimension, in our resource, as a function of the number of occurrences in the web (more than 5, 100, 1000 occurrences).}
\label{tbl:resource-filter-pos-counts}
\end{table}

%% file: eval.tex
\section{Evaluation Data}
\label{sec:eval}

In this section we describe the datasets we use for evaluation. For the dataset introduced in \cite{forbes2017verb}, we highlight a few problems we identified in it and how we corrected them, resulting in a new, cleaned up version of the dataset (Sec. \ref{sec:noun-eval}).
Moreover, since \rnacronym\ is more fine-grained than previous approaches supported, we also describe a new dataset for noun comparisons that was annotated by human annotators.
We then describe the evaluation used for comparing adjectives (Sec. \ref{sec:adj-eval}), and finally, an intrinsic evaluation done directly on the resource quality (Sec. \ref{sec:intrinsic-eval}).

\subsection{Commonsense Property Comparison}
\label{sec:noun-eval}

\citet{forbes2017verb} created a dataset consisting of 3,656 object pairs labeled by crowd workers.
The annotators were asked to label the typical relation between two objects along five dimensions: \textsc{size}, \textsc{weight}, \textsc{strength}, \textsc{rigidity} and \textsc{speed}: whether the first object was typically greater than, lesser than, or equal to the second along each dimension. $38$-$59$\% of the annotations (depending on the dimension) yielded perfect agreement among all annotators; $90$-$95$\% of them had an identifiable majority label, and they chose to keep all of these. We refer to this dataset as \fc.

\paragraph{Ill-Defined Comparisons}
In preliminary experiments on \fc\ we observed low results relative to the 76\% achieved by the current state-of-the-art \cite{yang2018extracting}. A close inspection of a sample of 100 pairs (20 from each dimension) revealed that only 57\% of the examples were in agreement with the annotations in \citet{forbes2017verb} and 47\% were not comparable. The most common reasons for disagreements were: (1) Broad objects: e.g. (\textit{father}, \textit{clothes}, \textit{big}); (2) Abstract objects: e.g. (\textit{seal}, \textit{place}, \textit{big}); (3) Ill-defined dimension: e.g. (\textit{friend}, \textit{bed}, \textit{strong}).

\paragraph{Training Leakage and New Split}
Another problem we identified in \fc, which results in a biased evaluation, is {\em leakage} from the training set to the dev/test sets. We identified two types of leakage. 

The first is a \textbf{Transitivity Leakage}: when the training set contains the tuples: ($o_1$, $o_2$, $d$) and ($o_2$, $o_3$, $d$), and the dev/test set contains the tuple ($o_1$, $o_3$, $d$). For example, the training set contains (`\textit{person}', `\textit{fox}', `\textit{weight}', `\textit{bigger}') and (`\textit{fox}', `\textit{goose}', `\textit{weight}', `\textit{bigger}'), and the dev set contains (`\textit{person}', `\textit{goose}', `\textit{weight}', `\textit{bigger}'). While transitivity is an inherent property of this data, success on the transitive closure of training examples does not reflect the ability of the algorithm to infer the correct relation between two unseen objects, and these examples should be removed from the evaluation data.
We found 4.3\% of the dev and 3.5\% of the test data had transitive leakage.

The second type of leakage we identified is \textbf{Object Leakage}, where a certain object in the dev/test set already appeared in the training set. This happens in 94.8\% and 95.7\% of the examples in the dev/test sets, respectively. This means that success on these objects might not reflect the generalization abilities of the algorithm, but rather a memorization of the training data. 

To address these concerns, we reorganized the train/dev/test sets, forming new splits, which we refer to as \fcl. The new split sizes can be found in Table \ref{tbl:splits}. We re-ran the current models on \fcl\ and, as expected, we observe a drop of 5-6\% in accuracy: from the original 76\% accuracy on the dev/test sets, to 70\% and 71\% accuracy, respectively.

\input{tables/FC_new_split.tex}

\paragraph{F\&C Re-annotation}
Due to the ill-defined comparison we identified in the dataset, we re-annotated it using crowd-source workers, who were trained with specific instructions to attend to the validity of the comparison.
We used 3 annotators per example and the majority vote was used as the final answer. Examples with no agreement, i.e., where each annotator chose a different option, were discarded. The inter annotator agreement yielded Fleiss kappa of $k=89.8$. Out of 7322 tuples in the original dataset, 59.5\% were discarded either because the objects were simply not comparable, or due to lack of agreement between the annotators. After removing the non-comparable examples, the kappa agreement was $k=97.2$. 
We refer to this new dataset as \fcc. We also tested the agreement between the new labels, and the corresponding labels in the original dataset, achieving near-perfect agreement of $k=90.2$, establishing the quality of the new annotations.

\paragraph{New, More Conservative Dataset}

Due to the problems we identified in \fc\ and the fact that it became significantly smaller after filtering out ill-defined comparisons, we created a new dataset. We provided human annotators with more precise 
definitions and restricted comparisons to specific domains using only a subset of the dimensions -- \textsc{mass}, \textsc{speed}, \textsc{currency} and \textsc{length}.
We further controlled the generation of comparable objects by using Category Builder (CB) \cite{MahabalRoMi18}, a method which can be used to expand a set of seed words into others in the same category. For each domain and dimension we fed an initial seed into CB, and used the top results as comparable pairs.
Table 3 
in the Appendix presents statistics and examples from each category from the new dataset.
Note that the new dataset is only used as a test set and thus \textit{leakage} is not applicable. Moreover, due to the controlled data generation process, we avoided some of the comparison issues we observed in \fc.

We used crowdsourcing to annotate the pairs, and obtained a substantial inter-annotator agreement of $k=77.1$. Each example was annotated by three annotators and we used majority vote to determine the final labels. The final dataset discards examples with no agreement and Non-Comparable label, resulting in 4,773 examples.

Our method for determining a relation between two objects is unsupervised and does not require a training set. However, in order to compare with other methods, we split it into train/dev/test of sizes 326/1101/1101 respectively, while keeping the split free of leakage. (Note that the number of training examples is comparable to \citet{yang2018extracting}.
This split contains only comparisons from the \textsc{speed}, \textsc{length} and \textsc{weight} dimensions for fair comparison to previous work.) 

\paragraph{The Relative Size Dataset}
\citet{bagherinezhad2016elephants} created a dataset of 486 object pairs between 41 physical objects. The dataset (which we refer to as \textsc{Relative}) focuses solely on the physical size dimension. Nevertheless, we use it as another evaluation of \rnacronym.

\subsection{Scalar Adjectives}
\label{sec:adj-eval}

\input{tables/adj_dataset_stats.tex}

Several test sets have been created to evaluate the intensity of adjectives. 
The dataset created by \citet{de2013good} uses adjective clusters based on the `dumbbell' structure of adjectives in WordNet e.g. ``\textit{cold} $<$ \textit{frigid} $<$ \textit{frozen}''.
\citet{wilkinson2016gold} created another testset, by defining a total order between adjectives in the same cluster, spanning the entire scale range.
For example, in the \textsc{size} domain, the full cluster is: ``minuscule $<$ \textit{tiny} $<$ \textit{small} $<$ \textit{big} $<$ \textit{large} $<$ \textit{huge} $<$ \textit{enormous} $<$ \textit{gigantic}''. A total of 60 adjectives were collected across 12 clusters.

Since our method only handles measurable objects, we manually removed all of the non-measurable clusters (e.g., ``\textit{known} $<$ \textit{famous} $<$ \textit{legendary}'' was removed) and evaluated on the rest. In this process we found that the new dataset by \citet{D18-1202} contains only a small number of measurable clusters and some overlap with the other testsets, therefore we exclude this test set from our evaluation.
The number of pair comparisons and unique objects are detailed in Table \ref{tbl:adj-splits}, both the original datasets and the subset we used in this work.

\subsection{Intrinsic Evaluation}
\label{sec:intrinsic-eval}
Lastly, since our resource is more expressive than what was done in this area before, we also conducted a novel intrinsic evaluation.
We ran the evaluation as follows: Given an object and a dimension, we extracted the median of the distribution, expanded it into a range and then asked human raters whether this range overlaps with the range of the target object-dimension pair. For example, when evaluating the speed of a car, its median is 99.7 km/h. We then convert it to a range of 10-100 km/h by relaxing it to its nearest order of magnitude numbers, and asked annotators if this range corresponds to the typical speed of a car.


We collected a total of 1,271 examples from the same pool of comparisons used for our new dataset.
Each example was evaluated by 3 annotators and labeled with the majority vote, discarding examples with no agreement.

%% file: tables/FC_new_split.tex
\begin{table}[t]
\resizebox{\columnwidth}{!}{%
\centering
\begin{tabular}{ll|rrr|r}
\multicolumn{2}{l}{}                      & Train    & Dev    & Test   & All       \\ \hline
\multirow{2}{*}{All Labels}        & \fc\  & 587       & 5,418  & 6,007  & 12,012    \\
                                   & \fcl\       & 712       & 3,000  & 4,497  & 8,209     \\ \hline
\multirow{2}{*}{Subset Labels}     & \fc\  & 361       & 3,311  & 3,650  & 7,322     \\
                                   & \fcc\     & 173       & 1,268  & 1,523  & 2,964
\end{tabular}
}
\caption{F\&C dataset size. All Labels represent the original dataset with all the labels. Subset Labels are the subset labels which are inferable by the resource.}
\label{tbl:splits}
\end{table}

%% file: tables/adj_dataset_stats.tex
\begin{table}[t]
\resizebox{\columnwidth}{!}{%
\begin{tabular}{ll|rrr}
                           &                & deMelo    & Wilk-intense & Wilk-all   \\ \hline
\multirow{2}{*}{All Labels}     & \#Objects & 300       & 60           & 60         \\
                                & \#Pairs   & 682       & 60           & 133        \\ \hline
\multirow{2}{*}{Subset Labels}  & \#Objects & 58        & 21           & 22         \\
                                & \#Pairs   & 121       & 21           & 47         \\     
\end{tabular}
}
\caption{Adjective dataset size from deMelo and Wilkinson. The All Labels row represent the original dataset, with all adjective clusters. The Subset Labels row describes the filtered partition on these datasets, with only measurable adjectives.}
\label{tbl:adj-splits}
\end{table}

%% file: results.tex
\section{Experimental Results}

The object comparison task described in Sec. \ref{sec:noun-eval} is formulated as: Given objects $o_1$ and $o_2$ and dimension $d$, predict the relation $y \in \{<, =, >\}$.\footnote{We use the three-way model described in \citet{forbes2017verb} and not its elaboration in \citet{yang2018extracting}.} To solve this task, we look up the set of measurements associated with each object-dimension pair in the object dictionary. For this evaluation, we aggregate all objects while ignoring their heads (as described in Sec.~\ref{sec:resource}). We compare the two distributions obtained by their medians. If the object-dimension pair does not appear in \rnacronym, we assign it a 0 value.

\input{tables/adj_algo.tex}
Adjective comparisons require a different treatment. Earlier work was done mainly on intensities, and so comparisons are only across half the scale (i.e., not on \textit{hot} vs. \textit{cold}, but on degrees of hot and degrees of cold separately). 
The dimension of the comparisons is not given explicitly; although it is possible to infer the most relevant dimension from \rnacronym\, it is not trivial and we leave this for future work. Instead, we manually label the dimension of each cluster used. For example, to the ``\textit{cold} $<$ \textit{frigid}'' comparison, we assign the \textsc{temperature} dimension.

The inference method for adjectives is also more subtle. As adjectives can describe a wide range of objects, their variance is higher than that of nouns. Therefore, our inference method makes use of an aggregation of individual objects: For each pair of adjectives we wish to compare, we query \rnacronym\ for every noun that both adjectives are seen to modify. For each such noun, we compare the distributions along the specified dimension, and choose the majority comparison over all such nouns as the prediction for the adjective pair. This process is outlined in Algorithm \ref{algo:adj-comparison}.

For the experiments using \rnacronym\ we used all three distance-based versions (sentence distance, 10 and 3 words distance). We found that the sentence-based has the higher coverage, but lower precision, whereas the lower distance-based has less coverage but higher precision.

\subsection{Comparative Evaluation}
\input{tables/noun_comp_results.tex}
\paragraph{Noun Comparison}
The left column of Table \ref{tbl:noun-comp-results} presents results for the cleaned version of the \citet{forbes2017verb} dataset. The current state-of-the-art model achieves a total accuracy on the test set of 87\%, while our best method achieves 80\%. First, we note that the accuracies are significantly higher than those on the original dataset, for all methods. Second, we still observe lower accuracy for our method compared to \citet{yang2018extracting}. We can attribute this gap to two reasons. First, they fine-tune their model on a training set, and although the training set size isn't large, it is necessary for achieving these results. 
Secondly, they are able to exploit similarities and capture synonym information through pre-trained word embeddings, which our method cannot. For example, the development set contains the comparison: (`\textit{lady}', `\textit{step}', `\textit{size}') and (`\textit{wife}', `\textit{ship}', `\textit{size}'). While these comparison are valid, they are less intuitive, and can be solved by embedding methods due to their proximity in the embedding space to similar words, such as `\textit{person}'. And indeed, when using the word `\textit{person}' in our method instead of `\textit{lady}' and `\textit{wife}', our method makes the correct prediction.\footnote{While it is possible to augment \rnacronym\ and access it via word embeddings, we chose not to do it in our experiments to better estimate the quality of the resource itself.}

Results on the new objects comparison dataset we created are shown in the rightmost column of Table \ref{tbl:noun-comp-results}. Although our method doesn't benefit from a split into train/dev/test, we split it nevertheless to compare to previous work. This split is created such that there is no leakage from the train to the dev/test sets. We get better results than previous methods on this dataset: 63\% and 61\% accuracy on the dev/test sets compared to 60\% and 57\%. These relatively low results on this new dataset indicate that it is more challenging.

\input{tables/Bagherinezhad_data.tex}
The last evaluation of noun comparatives is on \textsc{Relative} \cite{bagherinezhad2016elephants}, presented in Table \ref{tbl:Bagherinezhad-dataset}. We report the results of the original work, where the best score used a combination of visual and textual signals, achieving 83.5\% accuracy. We also tested the method by \citet{yang2018extracting} on this dataset. Since the dataset is small, we did not split it, and instead used the training set from \citet{forbes2017verb}. This can be viewed as a transfer learning evaluation. The accuracy achieved by this method is 85.8\%, surpassing the previous method by more than 2 points. We evaluated our method on this dataset, achieving a new state-of-the-art result of 87.7\% accuracy with $k=10$ as a filter method.

\subsection{Adjective Comparison}
For the scalar adjective datasets, we present an evaluation on the deMelo dataset \cite{de2013good}, and the Wilkinson dataset \cite{wilkinson2016gold}. Previous work is limited by the patterns used for extraction to comparing adjectives from the same half-cluster. As Wilkinson data contains the full scalar range, we also present results on the full range. We compare to \citet{de2013good}, using the re-implementation of \citet{D18-1202} for global ranking. We also use the new method of \citet{D18-1202} to evaluate. This work is not entirely comparable, as the coverage of the data depends on the exact method used i.e. the combination of patterns, lexicon-based evidence and paraphrasing. Therefore, for each dataset, we used the method that obtained the highest coverage. For pairs with no coverage, we chose random labels with uniform distribution. The method of \citet{de2013good} outperforms the rest for their dataset, while the method of \citet{D18-1202} performs best on the Wilkinson data. Our method does get comparable results on \citet{de2013good}, while on Wilkinson \cite{wilkinson2016gold} we lag behind by 9.1 points.
Finally, we do achieve good results when evaluating on the full range scale of Wilkinson - 89.1\% accuracy. All of the errors by our method on this dataset evaluation are indeed on the intensity level, and not between the extremes. We therefore conclude that our method is good at differentiating between the adjectives on the two tips of the scale.

In the Adjective comparison, we also observe the highest variance as a function of the context window size $k$.
While \rnacronym\ with $k=10$ achieves the best results on two of the three datasets, when $k=3$ the results suffer from a big drop in performance. We hypothesize that this performance gap is due to the higher variance in the use of adjectives vs. nouns, and our inference method that is based not on the adjective itself, but on all its modifying objects.


\input{tables/adj_intensity_results.tex}

\subsection{Intrinsic Evaluation}
We perform the following intrinsic evaluation to assess the distribution quality of the resource.
The results of the intrinsic evaluation on a sample of \rnacronym~are shown in Table \ref{tbl:intrinsic-eval}. The total agreement is 69\%, while the specific agreements for \textsc{mass}, \textsc{length}, \textsc{speed} and \textsc{currency} are 61\%, 79\%, 77\% and 58\% respectively. Originally, these annotations were performed by annotators from India and, while inspecting the annotation, we found cultural differences in the perceived prices of items. We re-annotated the samples in the currency category with annotators from the U.S. and found a much higher agreement score: 76\%. For example, Indian annotators reported that a suit could not cost between 1K-10K\$, while U.S-based annotators all reported it was possible.

\input{tables/intrinsic.tex}

%% file: tables/adj_algo.tex
\begin{algorithm}[ht]
    \caption{Adjectives Comparison Inference}
    \label{algo:adj-comparison}
    \begin{algorithmic}
        \setlength\parindent{1pt}
        \State \textbf{Input:}
        adjectives $x$,$z$, dimension $d$ and\\
        object distributions $H$
        \State \textbf{Output:}
        comparison label
        \State \textbf{Procedure:}
        \State Initialize $\hat{y}$, the predictions per head
         \State $intersect$ $\leftarrow$ findHeadIntersection($H$, $x$, $z$, $d$) \\\algorithmiccomment{the intersecting heads of $x$ and $z$}
         \For{ $a_i,b_i \in intersect$ }
         	\State $\hat{y_i}$ $\leftarrow$ compare($a_i$,$b_i$,$d$)
         \EndFor
        \State Return majority($\hat{y}$)
    
    \end{algorithmic}
\end{algorithm}

%% file: tables/noun_comp_results.tex
\begin{table}[t]
\begin{footnotesize}
\begin{tabular}{l|ll|ll}
Model/Dataset           & \multicolumn{2}{l}{F\&C Clean} & \multicolumn{2}{l}{New Data} \\
                        & Dev           & Test         & Dev           & Test         \\ \hline
Majority                & 0.54          & 0.57         & 0.51          & 0.50         \\
Yang et al. (PCE LSTM)  & \textbf{0.86}          & \bf{0.87}         & 0.60          & 0.57         \\ \hline
\rnacronym\               & 0.78          & 0.77         & \textbf{0.62}          & \textbf{0.62}         \\
\rnacronym\ + 10-distance & 0.78          & 0.77         & \textbf{0.62}          & \textbf{0.62}         \\
\rnacronym\ + 3-distance  & 0.81          & 0.80         & \textbf{0.62}          & 0.61        
\end{tabular}
\end{footnotesize}
\caption{Results on the noun comparison datasets.}
\label{tbl:noun-comp-results}
\end{table}

%% file: tables/Bagherinezhad_data.tex
\begin{table}[t]
\begin{footnotesize}
\centering
\begin{tabular}{l|r}
Model                               & Accuracy \\ \hline
Chance                              & 0.5      \\
Bagherinezhad et al.                & 0.835    \\ \hline
Yang et al. (Transfer)              & 0.858    \\ \hline
\rnacronym\                         & 0.872    \\
\rnacronym\ + 10-distance           & \textbf{0.877}    \\
\rnacronym\ + 3-distance            & 0.858       
\end{tabular}
\caption{Results on the \textsc{Relative} dataset. 
\citet{yang2018extracting} result was achieved by running their model on their training set, and using it as a transfer method on \textsc{Relative}. Finally, we present our own predictions, with different thresholds, which surpass previous work.}
\label{tbl:Bagherinezhad-dataset}
\end{footnotesize}
\end{table}

%% file: tables/adj_intensity_results.tex
\begin{table}[t]
\resizebox{\columnwidth}{!}{%
\begin{tabular}{l|l|l|l}
Model             & deMelo & Wilk-intense & Wilk-all \\ \hline
Global Ranking    & \textbf{0.642}  & 0.818  & - \\
Cocos et al.      & 0.620  & \textbf{0.841}  & - \\ \hline
\rnacronym         & 0.617  & 0.700  & 0.870 \\
\rnacronym\ + 10-distance     & 0.608  & 0.750  & \textbf{0.891} \\
\rnacronym\ + 3-distance      & 0.567  & 0.500  & 0.761

\end{tabular}
}
\caption{Results on scalar adjectives datasets.}
\label{tbl:adj-results}
\end{table}

%% file: tables/intrinsic.tex
\begin{table}[t]
\resizebox{\columnwidth}{!}{%
\begin{tabular}{l|rrrrr}
Method/Data      & Mass & Length & Speed & Currency & All \\ \hline
Indian Annotators    & 0.61 & 0.79   & 0.77  & 0.58     & 0.69 \\
US Annotators & -    & -      & -     & 0.76     & - 
\end{tabular}
}
\caption{Intrinsic Evaluation. Accuracy of the number of objects which our proposed median fall into range of the object, given the dimension.}
\label{tbl:intrinsic-eval}

\end{table}

%% file: discussion.tex
\section{Conclusion and Discussion}
\label{sec:discussion}



This paper develops an unsupervised method for collecting quantitative information from a large web corpus, and uses it to create \rnacronym, a very large resource consisting of distributions over physical quantities associated with nouns, adjectives, and events. We have evaluated \rnacronym~ on multiple existing and new datasets and showed that it compares favorably with other methods that require more resources and lack coverage relative to \rnacronym. Below, we discuss a few interesting issues brought up by the data collection process that should be addressed in future work.

\begin{figure}[t]
\centering
\subfloat[Measured average temperature]{
\includegraphics[width=0.22\textwidth]{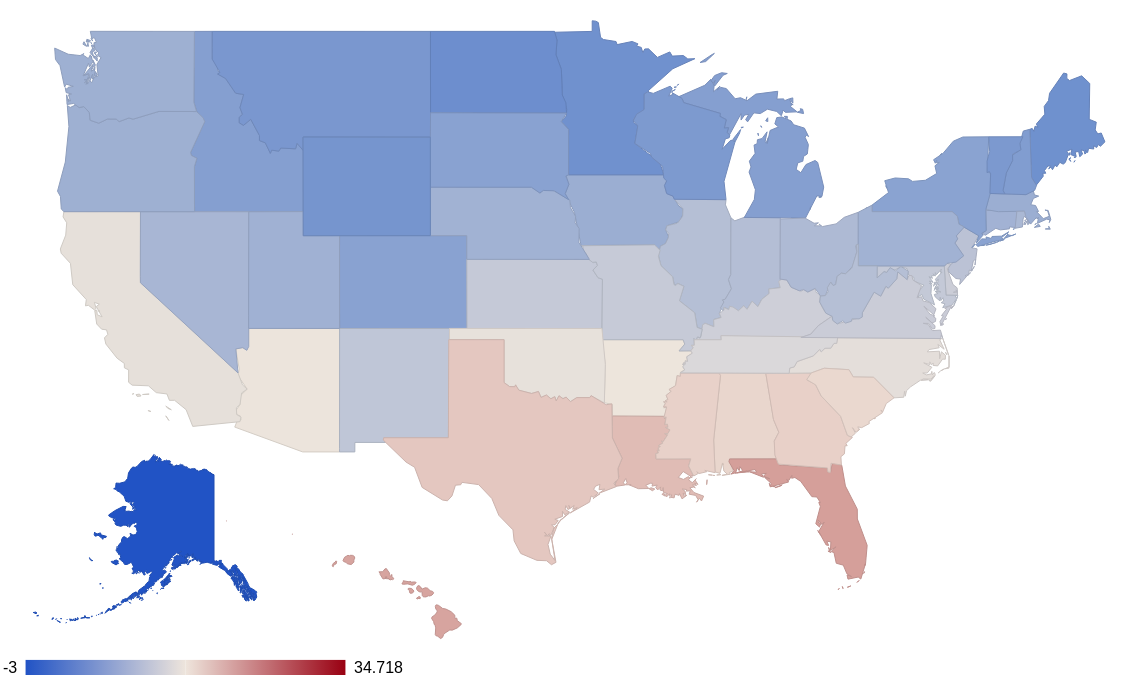}
\label{fig:sub-a-real-avg}
}~
\subfloat[Induced average temperature]{
\includegraphics[width=0.22\textwidth]{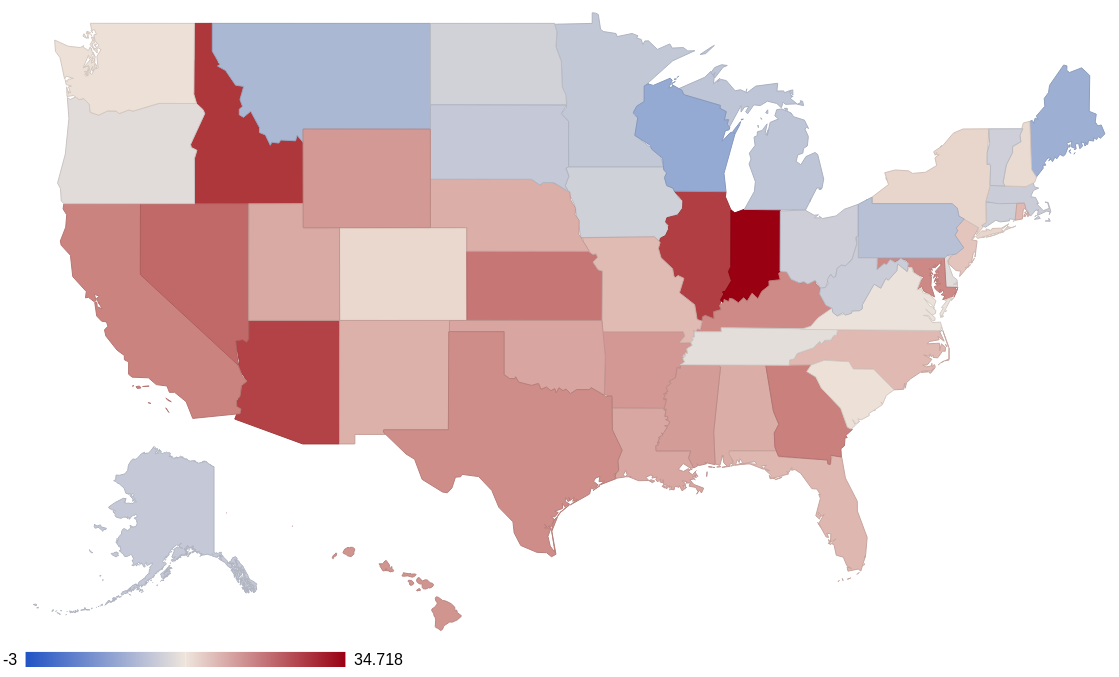}
\label{fig:sub-b-collected-avg}
}\\
\subfloat[Induced monthly temperature distributions]{
\includegraphics[width=0.45\textwidth]{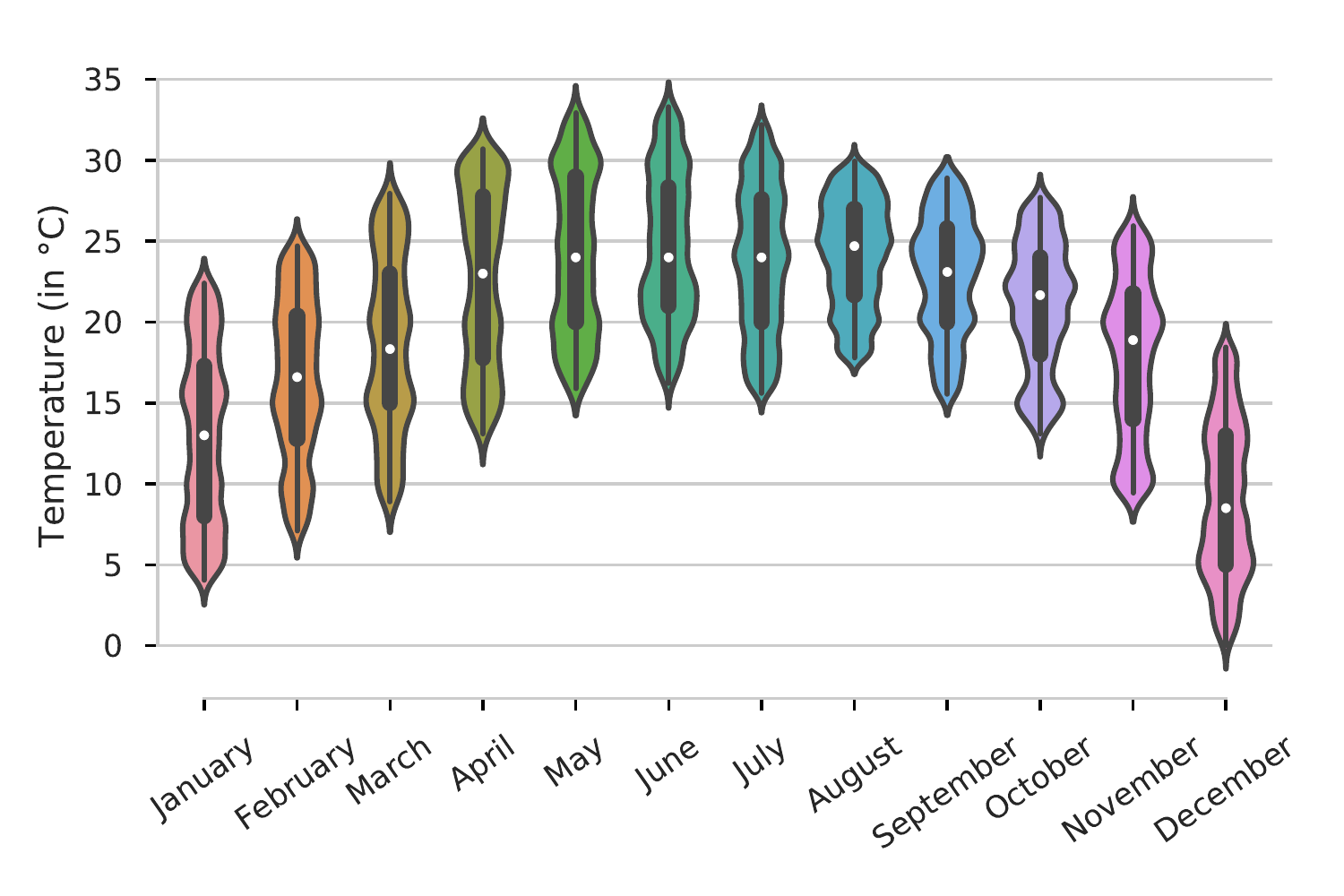}
\label{fig:months-temp}
}
\caption{Reporting bias: The observed average temperatures of U.S. states tend to extremes (Real average temperature (\ref{fig:sub-a-real-avg}) vs. the induced one (\ref{fig:sub-b-collected-avg})). Another sort of bias is exemplified in Figure \ref{fig:months-temp}, by a bias towards the northern hemisphere.}
\label{fig:med-temps}
\end{figure}

\paragraph{Reporting Bias and Exaggeration}
Although reporting bias \cite{gordon2013reporting} would seem to be a problem for a corpus-driven approach, in practice, DoQ is quite resilient to it due to the usage of very big web corpora and the collection method. As we do not rely on explicit comparisons between objects, but only on co-occurrences with numeric measurements, we can automatically infer relationships post-facto.
One form of reporting bias we  observe is that people are more likely to discuss objects when they are exceptional, or they exaggerate measurements for rhetorical effect, leading to long tails for some distributions (see ``\textit{slowest car}" in Figure \ref{fig:adj-comparison} and extreme temperatures in Figure \ref{fig:med-temps}). It is interesting to note that in the case of temperatures, both in the U.S states case (Figure \ref{fig:med-temps}) and the world case (Figure 2 in the Appendix), the exaggeration is towards hot temperatures, and not cold ones.

A somewhat different bias is shown in Figure \ref{fig:months-temp}; although the temperatures are an adequate representation of the cyclic year, it is highly biased towards the northern hemisphere, a result of the English web source data.

\begin{figure}[t]
\centering
\includegraphics[width=0.8\columnwidth]{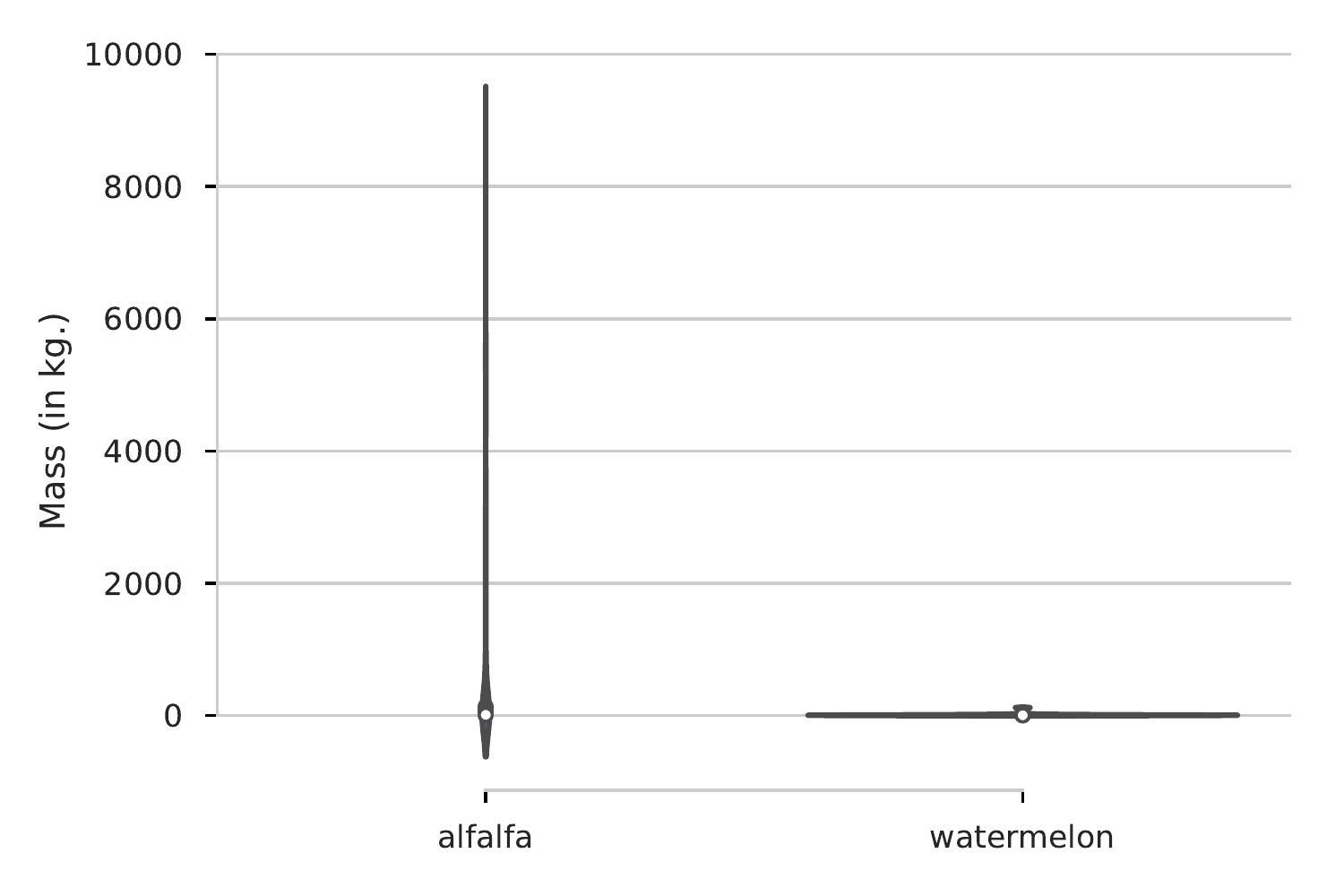}
\caption{Reporting bias can also be seen in this example where alfalfa's weight is induced in tons whereas in reality alfafa's weight is measure in grams.}
\label{fig:watermelon-alfalfa}
\end{figure}

A more subtle form of bias is due to attribution. For example, when comparing the size of alfalfa with the size of watermelons as shown in Figure \ref{fig:watermelon-alfalfa}, we see that alfalfa is mostly talked about in quantities in which it is harvested (order of tons) rather than individual units (grams).  This kind of bias cannot be identified as easily as the attribution bias discussed in Sec. \ref{sec:denoising}.

\paragraph{Polysemy}

We have not systematically explored how our resource performs on polysemous words and their senses, although our overall results indicate that in most cases the relatively biased distribution of polysemous senses render this a non-problem. We have also observed that in some cases the data itself can help disambiguate between different word senses.  For example, `bat' can refer to the animal, a baseball bat or a cricket bat. Figure 1 in the Appendix shows the induced distributions of length for these three senses of bat. While the distributions for ``Baseball bat'' (which measure about 1m) and ``Cricket bat'' (which may be no more than 956mm) are correct, the distribution for `bat' is probably a consolidation of these, the animal bat that can measure from 15cm to almost 1.7m in length, and some attribution noise (e.g. the distance the bat flew).

In conclusion, we developed and studied an unsupervised method for collecting quantitative information from large amounts of web data, and used it to create \rn\ (\rnacronym), a new, very large resource consisting of distributions over physical quantities associated with nouns, adjectives, and verbs.
The histogram version of the resource, as well as the new created dataset and evaluation code are available at \url{https://github.com/google-research-datasets/distribution-over-quantities}.

%% file: supplemental.tex
\section{Supplementary Material}
\label{sec:supplemental}

\begin{figure}[h!]
\centering
\includegraphics[width=1\columnwidth]{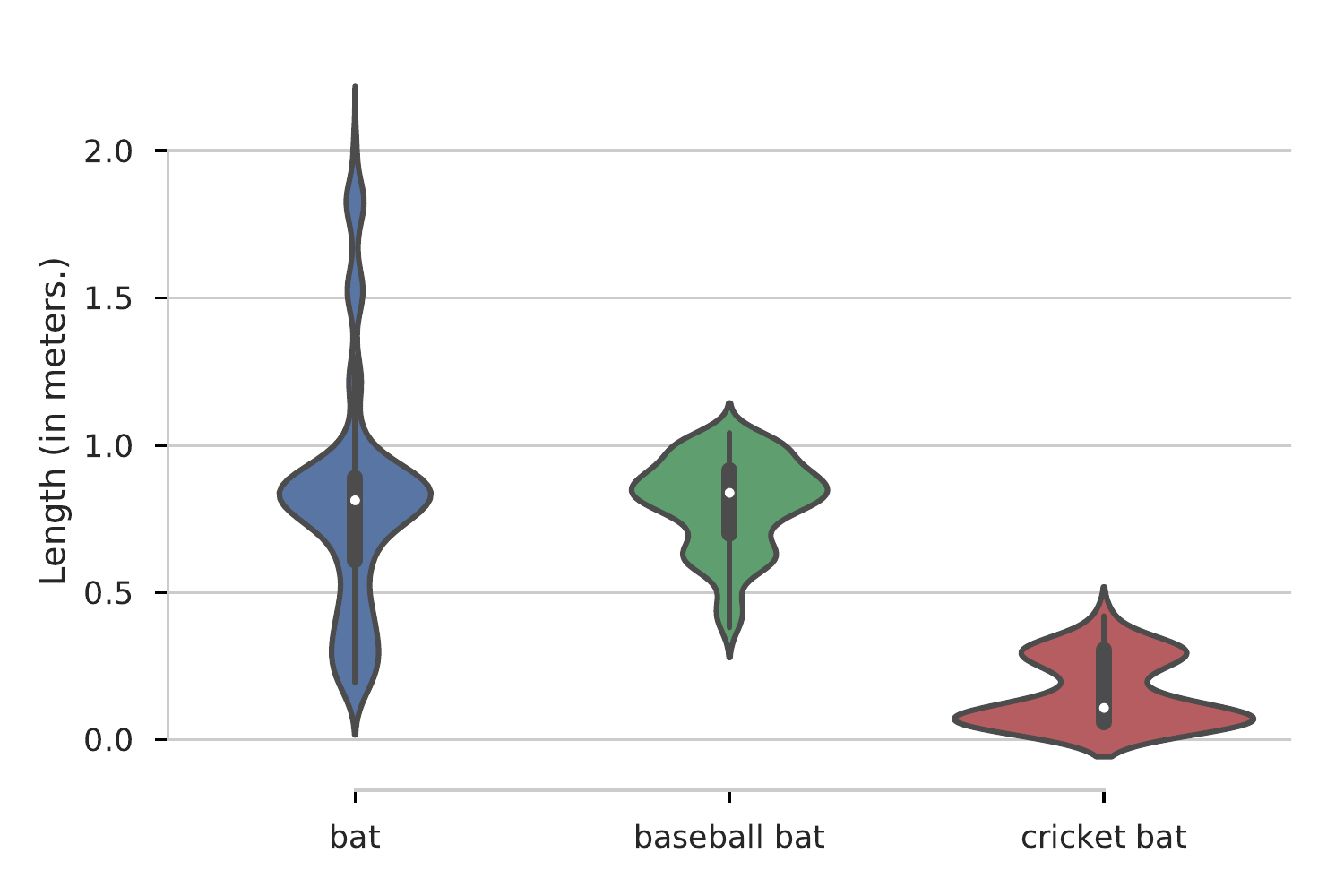}
\caption{Induced distributions for the various polysemous meanings of `bat'.}
\label{fig:bats}
\end{figure}

\begin{figure*}[ht]
\begin{multicols}{2}
    \subfloat[DoQ-induced average temperatures]{
    \includegraphics[width=1\columnwidth]{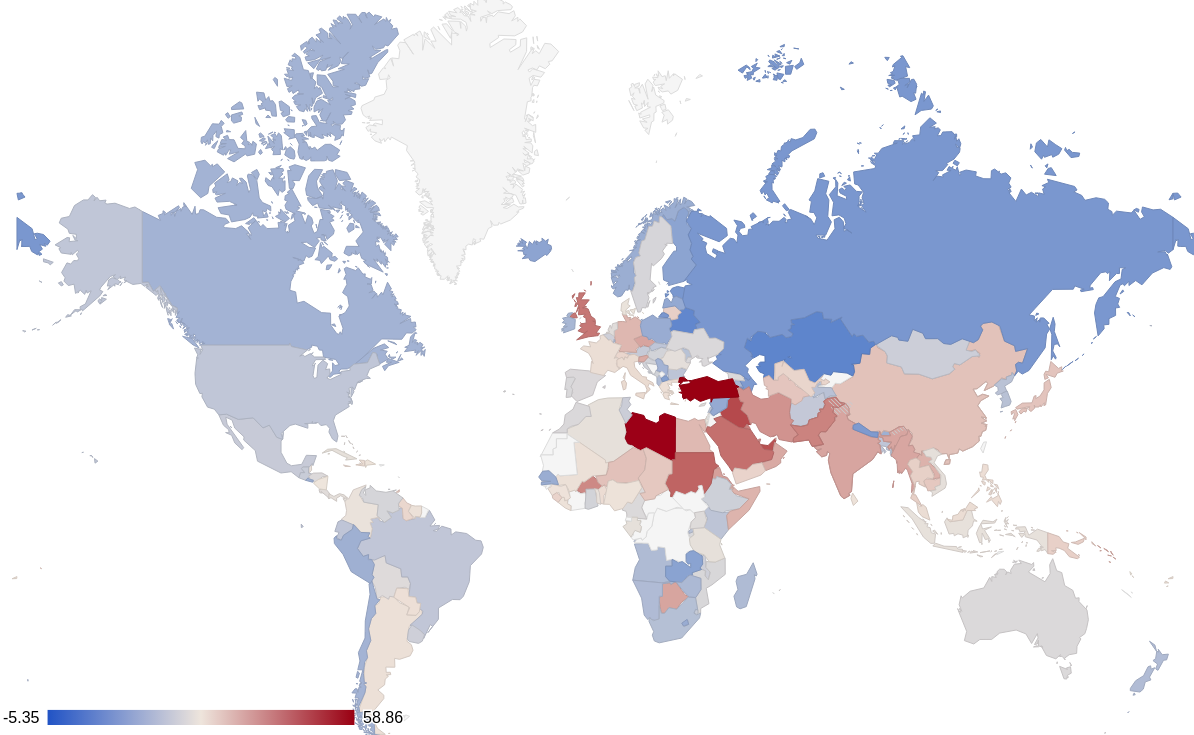}\par
    \label{lab:inf-world-temp}
    }
    \subfloat[Actual average temperature]{
    \includegraphics[width=1\columnwidth]{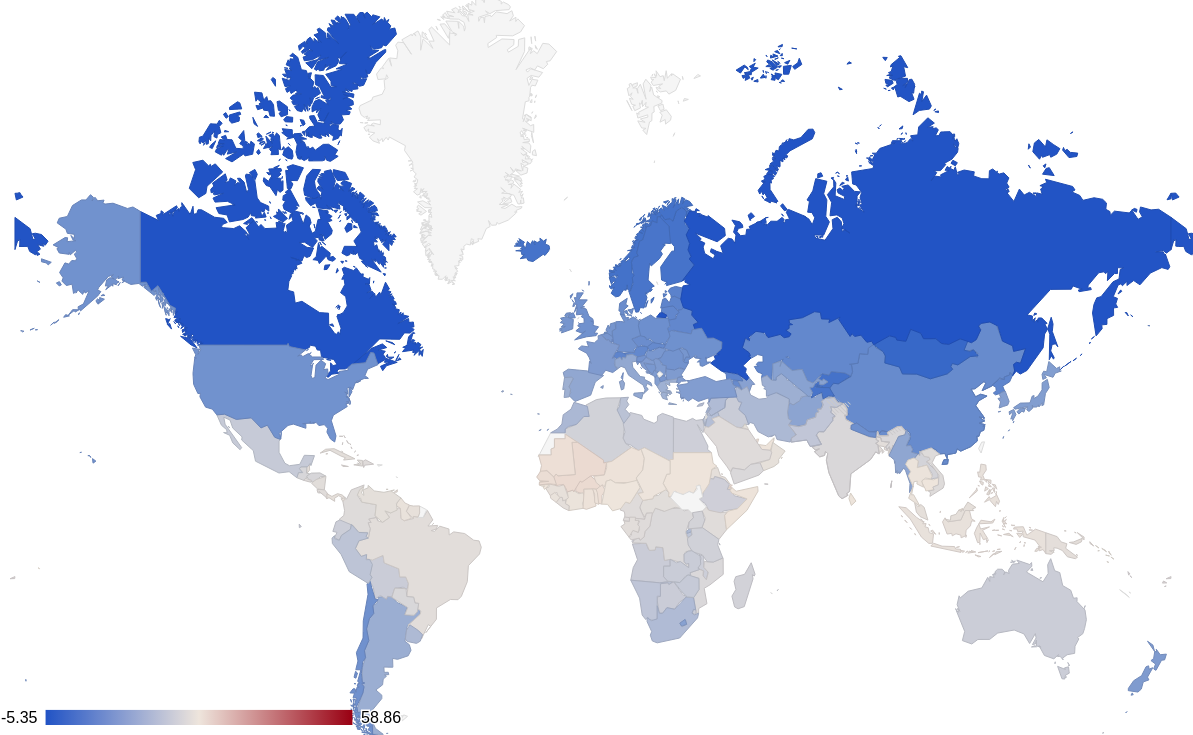}\par
    \label{lab:real-world-temp}
    }
    
    \end{multicols}

\caption{Average world temperatures, predicted by \rnacronym\ (Figure \ref{lab:inf-world-temp}), and the ground truth (Figure \ref{lab:real-world-temp}, taken from  \url{https://en.wikipedia.org/wiki/List_of_countries_by_average_yearly_temperature}). Interestingly, predicted temperatures are consistently higher than true ones, with the two hottest countries being Turkey (58.8\textdegree{}C) and Libya (57.7\textdegree{}C).}
\label{lab:world-temp}
\end{figure*}

\begin{figure*}[ht]
\centering
\subfloat[]{
\includegraphics[width=1\columnwidth]{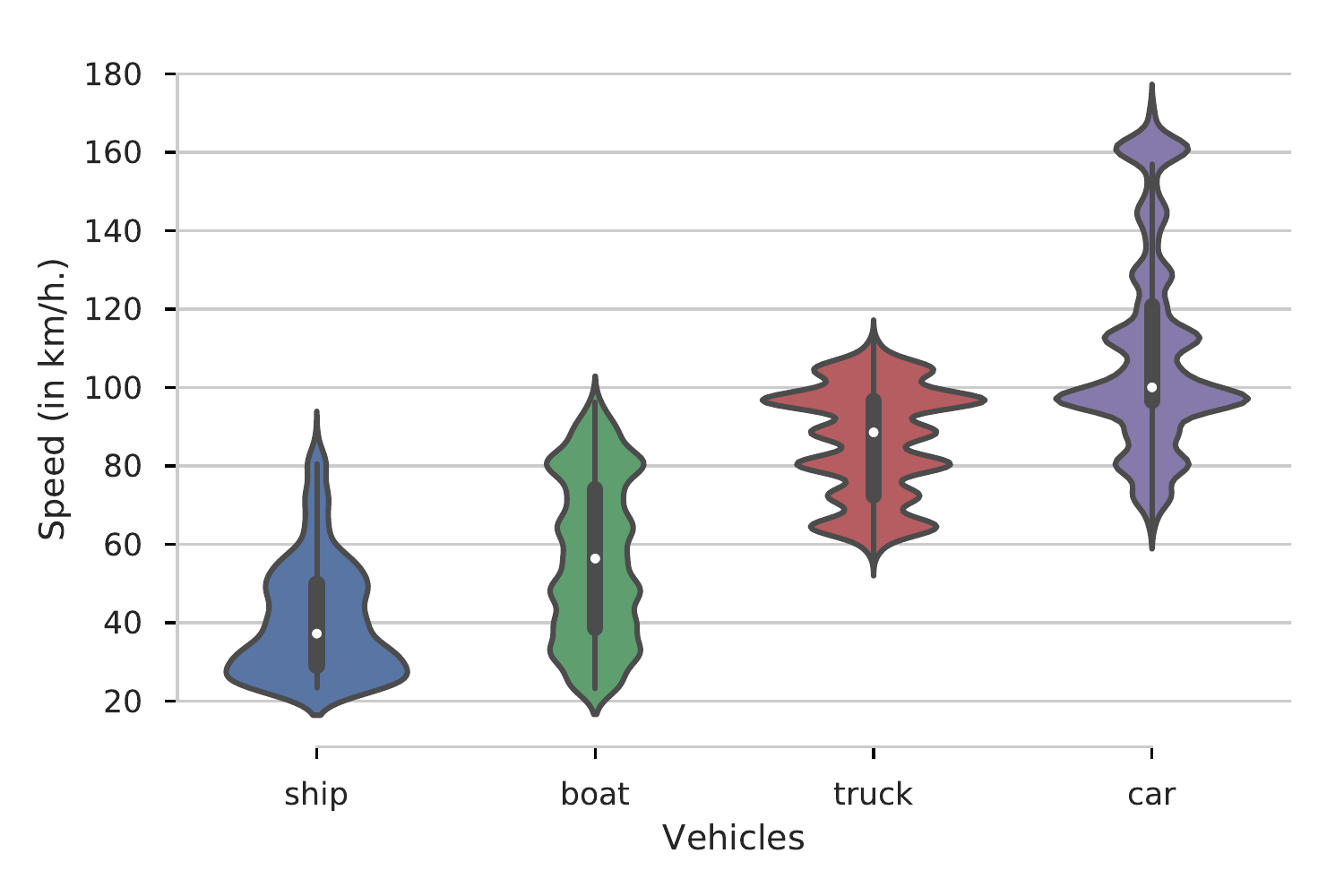}
\label{fig:vehicle_speed}
}
\subfloat[]{
\includegraphics[width=1\columnwidth]{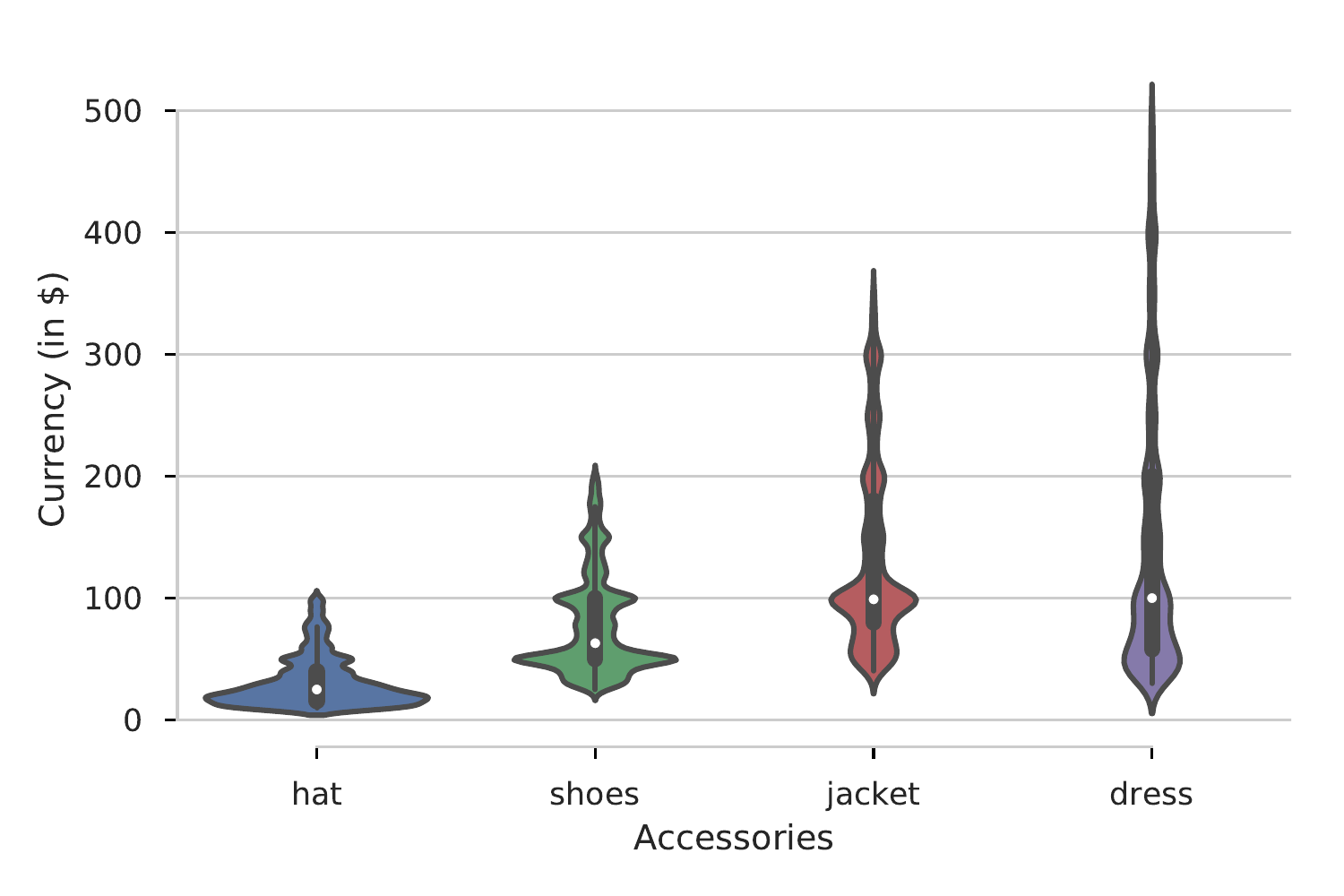}
\label{fig:accessories_dollar}
} \\
\subfloat[]{
\includegraphics[width=1\columnwidth]{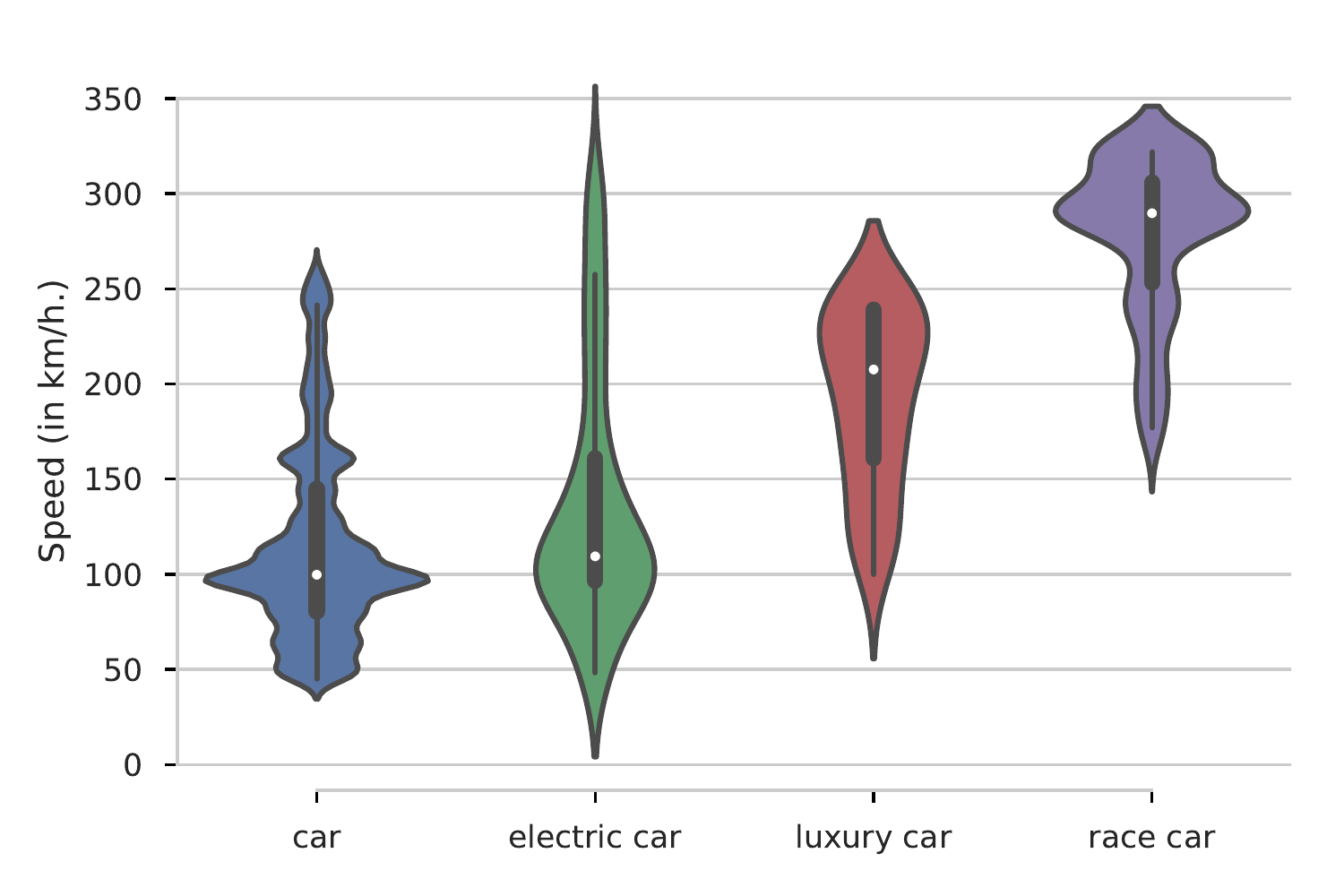}
\label{fig:car_speed}
}
\subfloat[]{
\includegraphics[width=1\columnwidth]{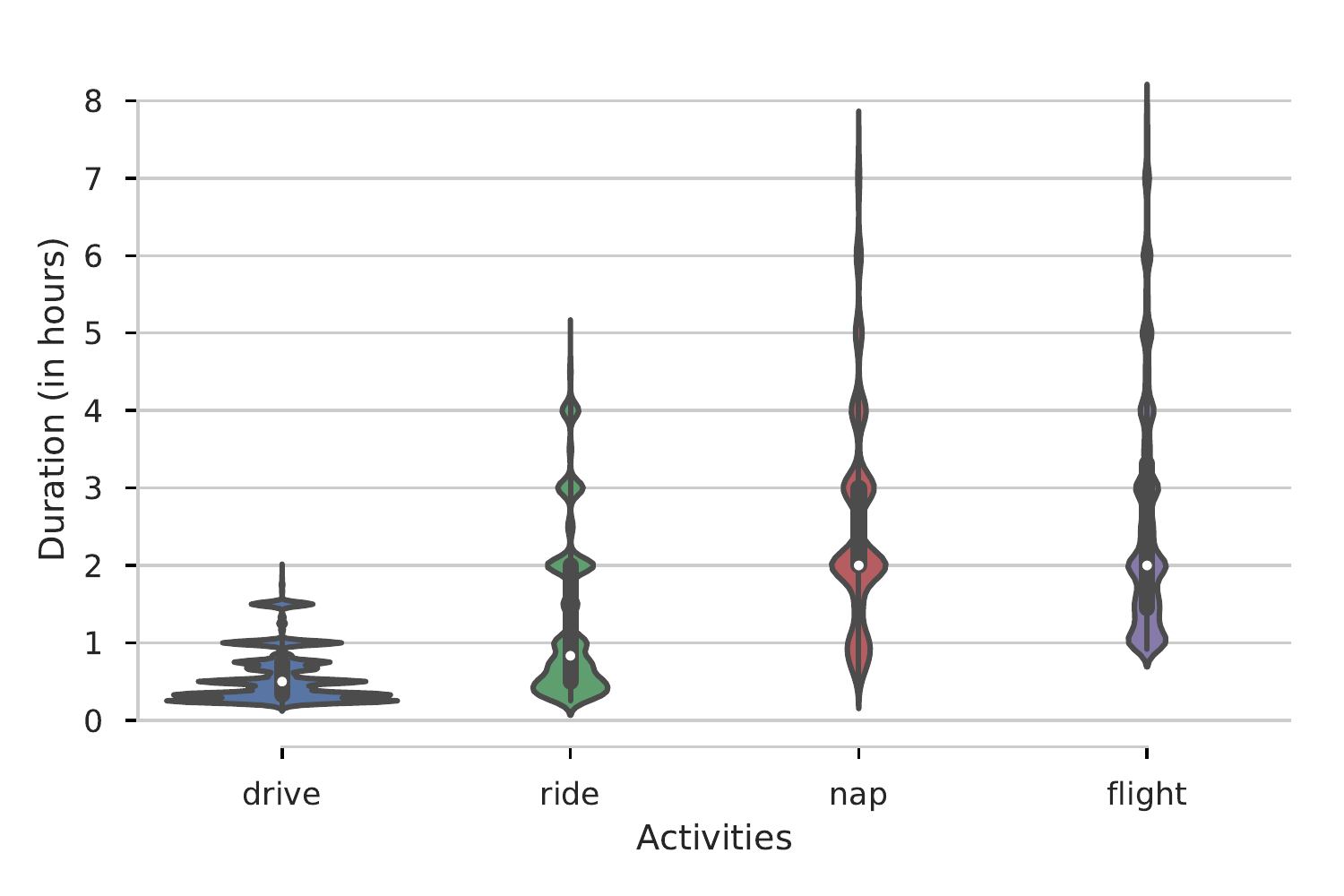}
\label{fig:activities_duration}
} \\
\subfloat[]{
\includegraphics[width=1\columnwidth]{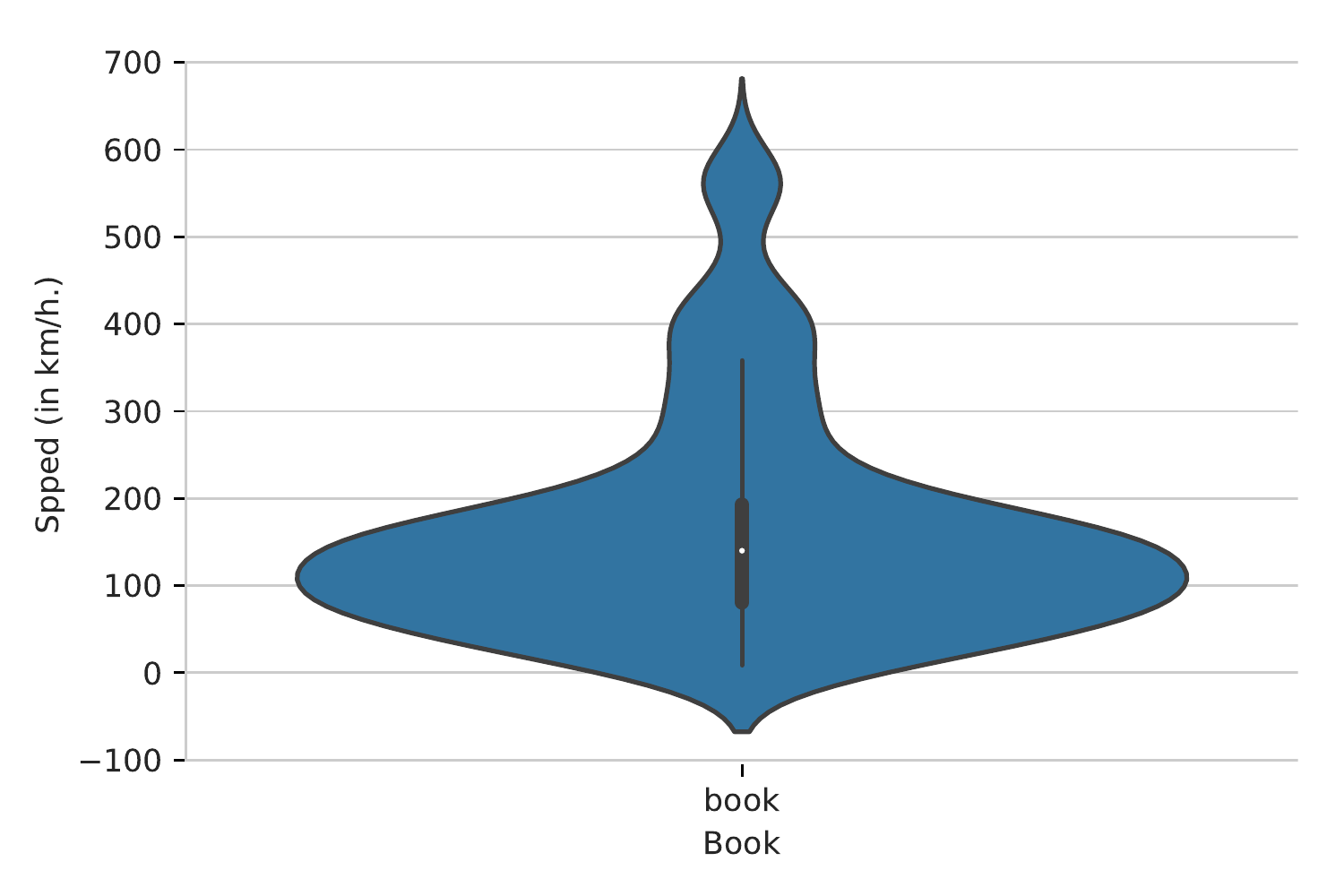}
\label{fig:book_speed}
}
\subfloat[]{
\includegraphics[width=1\columnwidth]{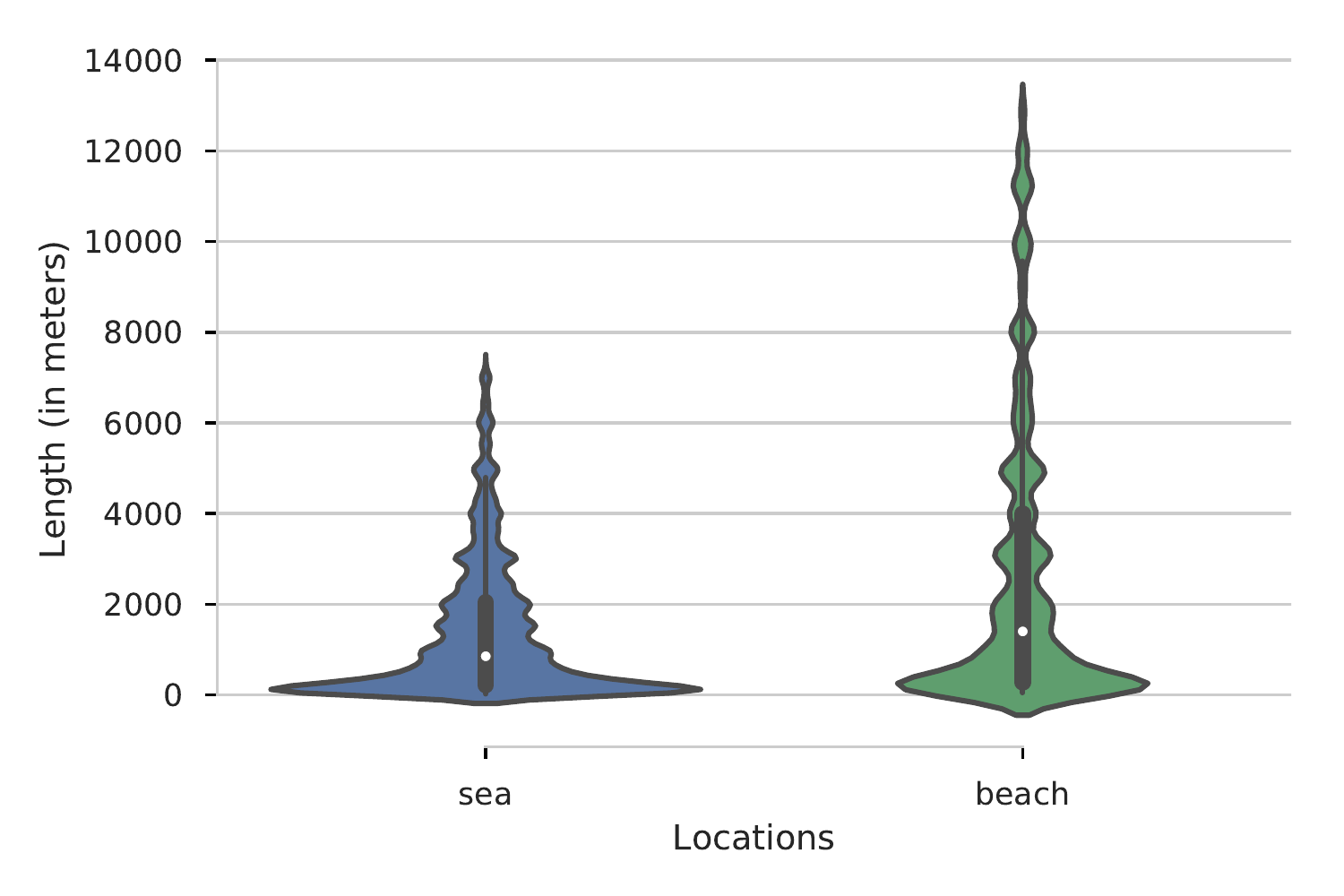}
\label{fig:attribution_err}
} \\

\caption{Comparisons of measurements extracted from DoQ for different categories of objects along their most salient dimensions. Note that (e) and (f) show spurious measurements for the speed of books (caused by attribution bias) and the lengths of seas and beaches (caused by confusion with sea levels).}
\label{fig:additional-measurements}
\end{figure*}

 In Figure \ref{fig:bats} we present length distributions for the word `bat' and two of its senses, showing a clear distinction between, what we believe to be, baseball bats and cricket bats. In Figure \ref{lab:world-temp}, we show the DoQ-induced and real average temperatures of countries.
Figure \ref{fig:additional-measurements} shows some additional measurement distributions from DoQ for various objects types and dimensions, as well as some examples of systematic errors caused by the collection/aggregation process.

In Table \ref{tbl:measurement-stats} we present specific example sentences used to extract correct measurements for objects along the ten most informative dimensions, while Table \ref{tbl:measurement-stats-bad} shows examples where incorrect measurement attributions were extracted.
Finally, Table \ref{tbl:objs-stats} describes the  different measurements and domains used in the new dataset we built, along with inter-annotator agreements, a representative example and the labeled relation.

\input{tables/measurements.tex}
\input{tables/measurements-bad-examples.tex}
\vspace{0.35in}
\input{tables/measure-domain-objects.tex}


%% file: tables/measurements.tex
\begin{table*}[t]
\begin{adjustbox}{width=1\textwidth}
\begin{tabular}{ll}
Dimension  & Example \\ \hline
TIME         & On 24 September drivers struck between 7.30 am and 8.30 am, the middle of the morning \textbf{rush hour}. \\
CURRENCY     & McKay picked up \$2,000 (the biggest winning \textbf{cheque} of her career) for winning the competition. \\
LENGTH       & The railway line has 140 bridges and 25 km of \textbf{tunnels} \\
AREA         & With over 36,000 acres, \textbf{Smoky Hill} is the Air National Guard's largest weapons range. \\
VOLUME       & The traditional Bordeaux barrique is an \textbf{oak barrel} with a capacity of 225 litres. \\
MASS         & The \textbf{men}'s 73 kg competition of the 2014 World Judo Championships was held on 27 August. \\
TEMPERATURE  & Places as far north as \textbf{New York City} reached 70 °F (21 °C) on Christmas Eve. \\
DURATION     & Formerly the \textbf{market} lasted for 14 days. \\
SPEED    &  Dora was the strongest \textbf{storm} of the year, peaking at 155 mph, just short of Category 5 status. \\
VOLTAGE      & The motorcycle used a 24 volt electric starter \textbf{motor} from a Douglas A-4B fighter plane. \\
\end{tabular}
\end{adjustbox}
\caption{Examples of sentences from the web corpus used to infer a correct measurement of the corresponding object for different dimensions.}
\label{tbl:measurement-stats}
\end{table*}

%% file: tables/measurements-bad-examples.tex
\begin{table*}[t]
\begin{adjustbox}{width=1\textwidth}
\begin{tabular}{ll}
Measurement & Example \\ \hline
TIME        & The largest \textbf{aftershock} measured 5.6 and struck nine minutes later at 1.51 pm. \\
CURRENCY    & Tuncay's move was confirmed the following \textbf{day}, with Stoke paying £5 million for the Turkish player. \\
LENGTH      & Everest from nine miles away in Rongbuk valley. \\
AREA        & The \textbf{plant} is spread over 25 acres of land. \\
VOLUME      & It was created for \textbf{cars} with engine size below 2 Litres. \\
MASS        & In recent years, the reported average annual \textbf{peanut} production lies around 828,000 tons (95\% for oil). \\
TEMPERATURE & The corresponding \textbf{phosphorine} and benzene undergo the analogous reaction at 100 °C and 200 °C, respectively. \\
DURATION    & Its \textbf{surface} remains frozen for around 155 days per year. \\
SPEED       & Both trains covered the 466 mile \textbf{route} at an average pace of 49 mph. \\
VOLTAGE     & The value of the shunt ensures a potential difference of 0.5 volt across it at the maximum \textbf{generator} load. \\

\end{tabular}
\end{adjustbox}
\caption{Examples of sentences from the web corpus from which an incorrect measurement was extracted, due to attribution/reporting bias or other reasons.}
\label{tbl:measurement-stats-bad}
\end{table*}

%% file: tables/measure-domain-objects.tex
\begin{table*}[t]

\begin{adjustbox}{width=1\textwidth}
\begin{tabular}{clrrlc}
\multicolumn{1}{l}{Measurement} & Domain              & Count & Inter-Annotator Agreement & Example & Relation         \\ \hline
\multirow{3}{*}{Mass}           & Vegtables \& Fruits &  500 &	76.9 &	(alfalfa, watermelon) & $<$     \\
                                & Animals             &  501 &	86.0 &	(skunk, lemur) & $<$            \\
                                & Technology          &  500 &	80.2 &	(modem, microphone) & $=$         \\ \hline
\multirow{3}{*}{Speed}          & Animals             &  500 &	83.3 &	(doll, dragon) & N.A.                \\
                                & Vehicles            &  500 &	78.8 &	(tractor, truck) & $<$          \\
                                & Activities          &  500 &	70.3 &	(barreling, traversing) & $>$    \\ \hline
\multirow{3}{*}{Currency}       & Dishes              &  500 &	74.3 &	(tiramisu, gnocchi) & $>$        \\
                                & Accessories         &  500 &	68.9 &	(bra, bikini) & $<$             \\
                                & Technology          &  500 &	76.2 &	(hose, sensor) & $>$             \\ \hline
\multirow{3}{*}{Length}         & Places              &  500 &	76.8 &	(bungalow, arena) & N.A.             \\
                                & House Object        &  500 &	72.6 &	(terrace, railing) & N.A.            \\
                                & Organs              &  499 &	73.2 &	(arm, ovary) & $>$               \\ \hline
All                             & All                 & 6,000&  77.1 & - & -        
\end{tabular}
\end{adjustbox}
\caption{Statistics for the new dataset, with the number of examples in each domain, their inter-annotator agreement, a representative example and the majority label assigned by annotators.}
\label{tbl:objs-stats}
\end{table*}

%% file: acl2019.bbl
\begin{thebibliography}{29}
\expandafter\ifx\csname natexlab\endcsname\relax\def\natexlab#1{#1}\fi

\bibitem[{Andor et~al.(2016)Andor, Alberti, Weiss, Severyn, Presta, Ganchev,
  Petrov, and Collins}]{andor2016globally}
Daniel Andor, Chris Alberti, David Weiss, Aliaksei Severyn, Alessandro Presta,
  Kuzman Ganchev, Slav Petrov, and Michael Collins. 2016.
\newblock Globally normalized transition-based neural networks.
\newblock In \emph{Proceedings of the 54th Annual Meeting of the Association
  for Computational Linguistics (Volume 1: Long Papers)}, volume~1, pages
  2442--2452.

\bibitem[{Bagherinezhad et~al.(2016)Bagherinezhad, Hajishirzi, Choi, and
  Farhadi}]{bagherinezhad2016elephants}
Hessam Bagherinezhad, Hannaneh Hajishirzi, Yejin Choi, and Ali Farhadi. 2016.
\newblock Are elephants bigger than butterflies? {R}easoning about sizes of
  objects.
\newblock In \emph{Thirtieth AAAI Conference on Artificial Intelligence}.

\bibitem[{Bollacker et~al.(2008)Bollacker, Evans, Paritosh, Sturge, and
  Taylor}]{bollacker2008freebase}
Kurt Bollacker, Colin Evans, Praveen Paritosh, Tim Sturge, and Jamie Taylor.
  2008.
\newblock Freebase: a collaboratively created graph database for structuring
  human knowledge.
\newblock In \emph{Proceedings of the 2008 ACM SIGMOD international conference
  on Management of data}, pages 1247--1250. AcM.

\bibitem[{Chaganty and Liang(2016)}]{chaganty2016much}
Arun Chaganty and Percy Liang. 2016.
\newblock How much is 131 million dollars? {P}utting numbers in perspective
  with compositional descriptions.
\newblock In \emph{Proceedings of the 54th Annual Meeting of the Association
  for Computational Linguistics (Volume 1: Long Papers)}, volume~1, pages
  578--587.

\bibitem[{Chambers et~al.(2010)Chambers, Raniwala, Perry, Adams, Henry,
  Bradshaw, and Weizenbaum}]{Chambers:2010:FEE:1806596.1806638}
Craig Chambers, Ashish Raniwala, Frances Perry, Stephen Adams, Robert~R. Henry,
  Robert Bradshaw, and Nathan Weizenbaum. 2010.
\newblock \href {https://doi.org/10.1145/1806596.1806638} {Flumejava: Easy,
  efficient data-parallel pipelines}.
\newblock In \emph{Proceedings of the 31st ACM SIGPLAN Conference on
  Programming Language Design and Implementation}, PLDI '10, pages 363--375,
  New York, NY, USA. ACM.

\bibitem[{Cocos et~al.(2018)Cocos, Wharton, Pavlick, Apidianaki, and
  Callison-Burch}]{D18-1202}
Anne Cocos, Veronica Wharton, Ellie Pavlick, Marianna Apidianaki, and Chris
  Callison-Burch. 2018.
\newblock \href {http://aclweb.org/anthology/D18-1202} {Learning scalar
  adjective intensity from paraphrases}.
\newblock In \emph{Proceedings of the 2018 Conference on Empirical Methods in
  Natural Language Processing}, pages 1752--1762. Association for Computational
  Linguistics.

\bibitem[{Dagan et~al.(2013)Dagan, Roth, Sammons, and Zanzotto}]{DRSZ13}
Ido Dagan, Dan Roth, Mark Sammons, and Fabio~Massimo Zanzotto. 2013.
\newblock Recognizing textual entailment: Models and applications.
\newblock \emph{Synthesis Lectures on Human Language Technologies},
  6(4):1--220.

\bibitem[{Dahlmeier and Ng(2011)}]{dahlmeier2011correcting}
Daniel Dahlmeier and Hwee~Tou Ng. 2011.
\newblock Correcting semantic collocation errors with l1-induced paraphrases.
\newblock In \emph{Proceedings of the Conference on Empirical Methods in
  Natural Language Processing}, pages 107--117. Association for Computational
  Linguistics.

\bibitem[{De~Melo and Bansal(2013)}]{de2013good}
Gerard De~Melo and Mohit Bansal. 2013.
\newblock Good, great, excellent: Global inference of semantic intensities.
\newblock \emph{Transactions of the Association for Computational Linguistics},
  1:279--290.

\bibitem[{Forbes and Choi(2017)}]{forbes2017verb}
Maxwell Forbes and Yejin Choi. 2017.
\newblock Verb physics: Relative physical knowledge of actions and objects.
\newblock In \emph{Proceedings of the 55th Annual Meeting of the Association
  for Computational Linguistics (Volume 1: Long Papers)}, pages 266--276.

\bibitem[{Goldberg and Orwant(2013)}]{goldberg2013dataset}
Yoav Goldberg and Jon Orwant. 2013.
\newblock A dataset of syntactic-ngrams over time from a very large corpus of
  english books.
\newblock \emph{Atlanta, Georgia, USA}, page 241.

\bibitem[{Gordon and Van~Durme(2013)}]{gordon2013reporting}
Jonathan Gordon and Benjamin Van~Durme. 2013.
\newblock Reporting bias and knowledge acquisition.
\newblock In \emph{Proceedings of the 2013 workshop on Automated knowledge base
  construction}, pages 25--30. ACM.

\bibitem[{Gusev et~al.(2011)Gusev, Chambers, Khaitan, Khilnani, Bethard, and
  Jurafsky}]{gusev2011using}
Andrey Gusev, Nathanael Chambers, Pranav Khaitan, Divye Khilnani, Steven
  Bethard, and Dan Jurafsky. 2011.
\newblock Using query patterns to learn the duration of events.
\newblock In \emph{Proceedings of the ninth international conference on
  computational semantics}, pages 145--154. Association for Computational
  Linguistics.

\bibitem[{Hearst(1992)}]{hearst1992automatic}
Marti~A Hearst. 1992.
\newblock Automatic acquisition of hyponyms from large text corpora.
\newblock In \emph{Proceedings of the 14th conference on Computational
  linguistics-Volume 2}, pages 539--545. Association for Computational
  Linguistics.

\bibitem[{Kim and de~Marneffe(2013)}]{kim2013deriving}
Joo-Kyung Kim and Marie-Catherine de~Marneffe. 2013.
\newblock Deriving adjectival scales from continuous space word
  representations.
\newblock In \emph{Proceedings of the 2013 Conference on Empirical Methods in
  Natural Language Processing}, pages 1625--1630.

\bibitem[{Kozareva and Hovy(2011)}]{kozareva2011learning}
Zornitsa Kozareva and Eduard Hovy. 2011.
\newblock Learning temporal information for states and events.
\newblock In \emph{2011 IEEE Fifth International Conference on Semantic
  Computing}, pages 424--429. IEEE.

\bibitem[{Lin et~al.(2012)Lin, Michel, Aiden, Orwant, Brockman, and
  Petrov}]{lin2012syntactic}
Yuri Lin, Jean-Baptiste Michel, Erez~Lieberman Aiden, Jon Orwant, Will
  Brockman, and Slav Petrov. 2012.
\newblock Syntactic annotations for the google books ngram corpus.
\newblock In \emph{Proceedings of the ACL 2012 system demonstrations}, pages
  169--174. Association for Computational Linguistics.

\bibitem[{Mahabal et~al.(2018)Mahabal, Roth, and Mittal}]{MahabalRoMi18}
Abhijit~A. Mahabal, Dan Roth, and Sid Mittal. 2018.
\newblock \href {http://cogcomp.org/papers/MahabalRoMi18.pdf} {Robust handling
  of polysemy via sparse representations}.
\newblock In \emph{*SEM}.

\bibitem[{Narisawa et~al.(2013)Narisawa, Watanabe, Mizuno, Okazaki, and
  Inui}]{narisawa2013204}
Katsuma Narisawa, Yotaro Watanabe, Junta Mizuno, Naoaki Okazaki, and Kentaro
  Inui. 2013.
\newblock Is a 204 cm man tall or small? {A}cquisition of numerical common
  sense from the web.
\newblock In \emph{Proceedings of the 51st Annual Meeting of the Association
  for Computational Linguistics (Volume 1: Long Papers)}, volume~1, pages
  382--391.

\bibitem[{Ning et~al.(2018)Ning, Wu, Peng, and Roth}]{NWPR18}
Qiang Ning, Hao Wu, Haoruo Peng, and Dan Roth. 2018.
\newblock \href {http://cogcomp.org/papers/NingWuPeRo18.pdf} {Improving
  temporal relation extraction with a globally acquired statistical resource}.
\newblock In \emph{Proceedings of the 2018 North American Chapter of the
  Association for Computational Linguistics (NAACL2018)}, pages 841--851.

\bibitem[{Pan et~al.(2006)Pan, Mulkar, and Hobbs}]{pan2006annotated}
Feng Pan, Rutu Mulkar, and Jerry~R Hobbs. 2006.
\newblock An annotated corpus of typical durations of events.
\newblock In \emph{LREC}, pages 77--82. Citeseer.

\bibitem[{Shivade et~al.(2015)Shivade, de~Marneffe, Fosler-Lussier, and
  Lai}]{shivade2015corpus}
Chaitanya Shivade, Marie-Catherine de~Marneffe, Eric Fosler-Lussier, and
  Albert~M Lai. 2015.
\newblock Corpus-based discovery of semantic intensity scales.
\newblock In \emph{Proceedings of the 2015 Conference of the North American
  Chapter of the Association for Computational Linguistics: Human Language
  Technologies}, pages 483--493.

\bibitem[{Shivade et~al.(2016)Shivade, de~Marneffe, Fosler-Lussier, and
  Lai}]{shivade2016identification}
Chaitanya Shivade, Marie-Catherine de~Marneffe, Eric Fosler-Lussier, and
  Albert~M Lai. 2016.
\newblock Identification, characterization, and grounding of gradable terms in
  clinical text.
\newblock In \emph{Proceedings of the 15th Workshop on Biomedical Natural
  Language Processing}, pages 17--26.

\bibitem[{Spithourakis et~al.(2016)Spithourakis, Augenstein, and
  Riedel}]{spithourakis2016numerically}
Georgios Spithourakis, Isabelle Augenstein, and Sebastian Riedel. 2016.
\newblock Numerically grounded language models for semantic error correction.
\newblock In \emph{Proceedings of the 2016 Conference on Empirical Methods in
  Natural Language Processing}, pages 987--992.

\bibitem[{Spithourakis and Riedel(2018)}]{spithourakis2018numeracy}
Georgios Spithourakis and Sebastian Riedel. 2018.
\newblock Numeracy for language models: Evaluating and improving their ability
  to predict numbers.
\newblock In \emph{Proceedings of the 56th Annual Meeting of the Association
  for Computational Linguistics (Volume 1: Long Papers)}, pages 2104--2115.

\bibitem[{Tandon et~al.(2014)Tandon, De~Melo, and Weikum}]{tandon2014acquiring}
Niket Tandon, Gerard De~Melo, and Gerhard Weikum. 2014.
\newblock Acquiring comparative commonsense knowledge from the web.
\newblock In \emph{Twenty-Eighth AAAI Conference on Artificial Intelligence}.

\bibitem[{Wilkinson and Oates(2016)}]{wilkinson2016gold}
Bryan Wilkinson and Tim Oates. 2016.
\newblock A gold standard for scalar adjectives.
\newblock In \emph{LREC}.

\bibitem[{Winn and Muresan(2018)}]{winn2018lighter}
Olivia Winn and Smaranda Muresan. 2018.
\newblock {’Lighter’} can still be dark: Modeling comparative color
  descriptions.
\newblock In \emph{Proceedings of the 56th Annual Meeting of the Association
  for Computational Linguistics (Volume 2: Short Papers)}, volume~2, pages
  790--795.

\bibitem[{Yang et~al.(2018)Yang, Birnbaum, Wang, and
  Downey}]{yang2018extracting}
Yiben Yang, Larry Birnbaum, Ji-Ping Wang, and Doug Downey. 2018.
\newblock Extracting commonsense properties from embeddings with limited human
  guidance.
\newblock In \emph{Proceedings of the 56th Annual Meeting of the Association
  for Computational Linguistics (Volume 2: Short Papers)}, volume~2, pages
  644--649.

\end{thebibliography}
